\documentclass[10pt,twocolumn,letterpaper]{article}

\usepackage{cvpr}              
\usepackage[table]{xcolor}
\definecolor{lightgray}{gray}{0.93} 
\definecolor{cvprblue}{rgb}{0.21,0.49,0.74}
\usepackage{multirow} 
\usepackage[pagebackref,breaklinks,colorlinks,allcolors=cvprblue]{hyperref}

\def\paperID{2805} 
\def\confName{CVPR}
\def\confYear{2026}

\title{UniD-Shift: Towards Unified Semantic Segmentation via Interpretable Shared–Private Multimodal Decomposition}

\author{
Shuai~Zhang$^{1}$, 
Zhecheng~Shi$^{1}$, 
Zhuoxiao~Li$^{1}$, 
Jing~Ou$^{1}$, 
Tengxi~Wang$^{1}$, 
Yuan~Liu$^{2}$, 
Wufan~Zhao$^{1*}$\\
$^{1}$HKUST(GZ), $^{2}$HKUST\\
{\tt\small
\{szhang240, jou719, twang744\}@connect.hkust-gz.edu.cn, 
yuanly@ust.hk}\\
{\tt\small
\{zhechengs, zhuoxiaoli, wufanzhao\}@hkust-gz.edu.cn
}
}

\begin{document}
\maketitle
\begin{abstract}

Semantic segmentation of large-scale 3D point clouds is crucial for applications such as autonomous driving and urban digital twins. However, the sparse sampling pattern of LiDAR and the view-dependent geometric distortion in image observations complicate cross-modal alignment and hinder stable fusion. Inspired by the fact that 2D images captured by cameras are representations of the 3D world, we recognize that the features learned from 2D and 3D segmentation share some common semantics, while other aspects remain modality-specific. This insight motivates a unified multimodal framework for joint 2D-3D semantic segmentation.
We combine a SAM-based vision encoder with a SPTNet-based geometric encoder to extract complementary semantic and geometric representations. The resulting features from both modalities are explicitly decomposed into shared and private subspaces, where the shared components summarize semantic factors common to both domains, and the private components preserve properties that are unique to each modality. A lightweight attention-based fusion module aggregates the shared features into a consistent cross-modal representation, and a regularized training objective ensures both semantic alignment and subspace independence. Experiments on the SemanticKITTI and nuScenes benchmarks demonstrate consistent improvements in segmentation accuracy over representative multimodal baselines, accompanied by competitive computational efficiency. Cross-domain evaluation on nuScenes USA-Singapore shows stable performance under distribution shifts, demonstrating strong generalization. The implementation code is publicly available at: \href{https://github.com/shuaizhang69/UniD-Shift}{https://github.com/shuaizhang69/UniD-Shift.}

\end{abstract}

\section{Introduction}
\label{sec:intro}

\begin{figure}[htbp]
    \centering
    \includegraphics[width=0.9\linewidth]{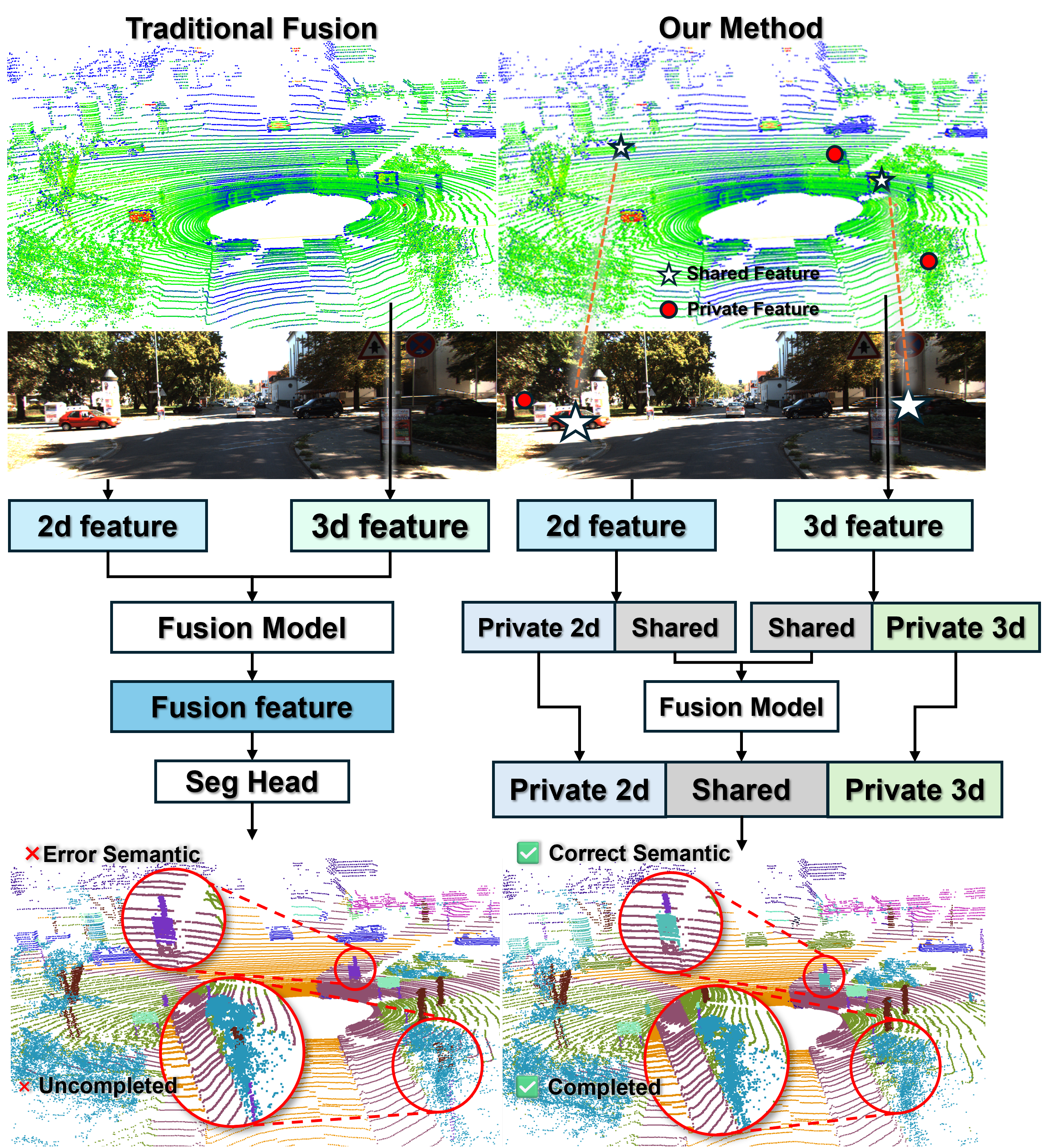}
    \caption{Existing multimodal fusion strategies (left) often mix 2D and 3D features without explicit semantic disentanglement, leading to redundant and ambiguous representations. Our proposed shared-private fusion framework (right) separates modality-invariant and modality-specific features, enabling efficient and interpretable 2D-3D interaction.}
    \label{fig:starting}
    \vspace{-12pt}
\end{figure}

Semantic segmentation of large-scale 3D point clouds is a central problem in intelligent perception and supports key applications in autonomous driving, urban reconstruction, and digital-twin modeling \cite{qu2025end, zhang2024point, zhao2025bfanet, guo2020deep, kolodiazhnyi2024oneformer3d}. Assigning semantic labels to individual points produces a structured representation of urban environments and enables reliable analysis of complex outdoor scenes. Current sensing systems often combine LiDAR measurements with RGB images to capture both geometry and appearance. LiDAR provides accurate spatial structure, whereas images offer dense visual information that reflects surface color and texture. Although these observations originate from different sensing principles, they describe the same physical world and share consistent semantic content such as object categories and spatial relations. Their complementary characteristics indicate that integrating 2D and 3D observations is an effective strategy for improving the robustness and completeness of scene understanding.

Recent progress in multimodal scene understanding has significantly advanced 3D semantic segmentation within single-domain scenarios. As illustrated in Fig.~\ref{fig:starting}, existing 2D-3D fusion methods consistently demonstrate that combining geometric structure from LiDAR with appearance information from images improves feature quality and produces clearer semantic boundaries \cite{jaritz2022cross, luo2025csfnet,luo2025paseg,xu2021rpvnet}. Fusion architectures that incorporate cross-modal alignment and joint reasoning further enhance the consistency of geometric and visual representations.Despite these advances, many approaches rely on heavy feature concatenation or large attention modules, increasing computational cost and limiting scalability in large outdoor scenes. Their reliance on full-feature fusion often introduces redundant representations and leads to unstable optimization during training. These limitations reveal a persistent gap between accurate multimodal integration and stable model behavior, indicating the need for a fusion strategy that maintains efficiency while preserving interpretability.

To address these limitations, we introduce a shared-private feature decomposition framework called \textbf{UniD-Shift}. The central assumption is that 2D and 3D modalities contain both common attributes reflecting consistent semantics and unique attributes encoding modality-dependent characteristics. Effective multimodal learning therefore requires integrating shared attributes while preserving distinctive modality-specific information. To achieve this, the proposed decomposition separates modality-invariant representations that describe stable semantic concepts from modality-specific representations that retain appearance variation in images and geometric detail in point clouds. This separation aligns with the inherent distinction between image-based appearance and LiDAR geometry.  By modeling shared and modality-specific information separately, the framework promotes consistent semantic correspondence while reducing interference from irrelevant modality-dependent variations. The resulting representation is more compact and stable, improving training stability and cross-domain generalization.

Overall, our contributions are summarized as follows:
\begin{itemize}
    \item A unified multimodal framework is proposed to achieve both high single-domain accuracy and strong cross-domain generalization through  feature interaction.  
    \item A shared-private decomposition mechanism enables explicit disentanglement of modality-invariant and modality-specific representations, improving semantic alignment and interpretability.  
    \item A dual-branch design integrating a SAM-based vision encoder with a sparse convolution-transformer backbone effectively combines semantic richness and geometric precision.  
    \item Extensive experiments on nuScenes and SemanticKITTI demonstrate superior segmentation accuracy and robustness compared with existing 2D-3D fusion and domain adaptation methods.
\end{itemize}

\section{Related work}
\label{sec:formatting}

\subsection{LiDAR-based 3D Semantic Segmentation}


\begin{figure*}
    \centering
    \vspace{-8pt}
    \includegraphics[width=0.8\linewidth]{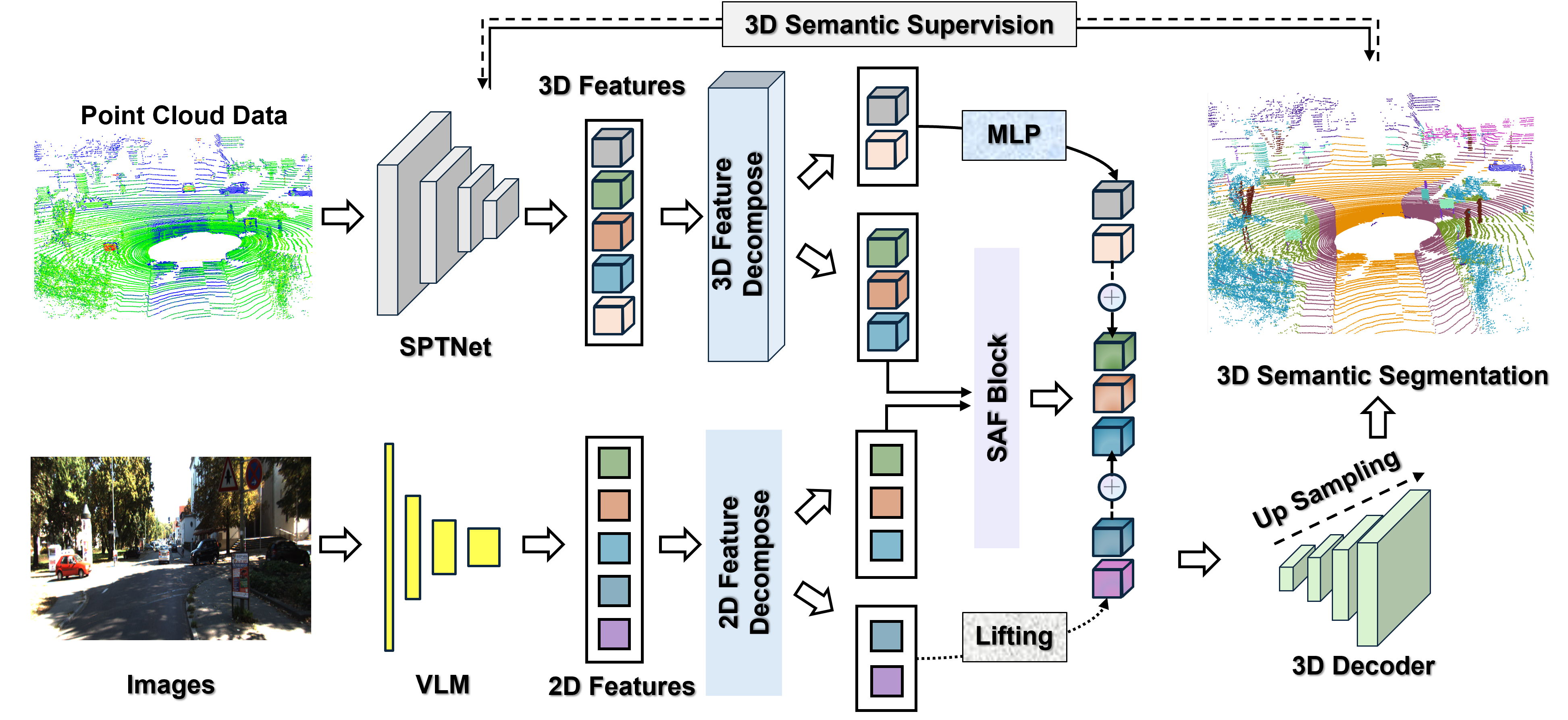}
    \caption{Overall architecture of the proposed \textbf{UniD-Shift}. 
    The model takes synchronized 2D images and 3D point clouds as inputs. 
    The 3D branch employs a  SPTNet backbone to extract hierarchical geometric features,  while the 2D branch utilizes a SAM-based encoder to obtain semantically enriched visual representations. 
    Both modalities are decomposed into shared and private components, then feed into the Shared Attention Fusion (SAF) block into a unified latent representation.}
    \label{fig:framework}
    \vspace{-12pt}
\end{figure*}

Existing methods for LiDAR-based 3D semantic segmentation are generally categorized into point-based~\cite{qi2017pointnet,qi2017pointnet++,thomas2019kpconv}, projection-based~\cite{deng2021multi,hamdimvtn,liang2018deep, su2015multi,yang2018pixor}, and voxel-based frameworks~\cite{choy20194d,graham20183d,hou20193d,jiang2020pointgroup,lang2019pointpillars,maturana2015voxnet,song2016deep,yan2018second,zhou2021panoptic}  .
Point-based models directly process irregular point sets to preserve fine geometric structures, yet their computational overhead limits scalability in large outdoor scenes.
Projection-based approaches transform point clouds into structured range representations, enabling efficient 2D convolutional or transformer-based processing but introducing spatial distortions and information loss in occluded regions.
Voxel-based methods discretize the 3D space into regular grids and use sparse convolutions to model volumetric structure. This strategy offers a balanced compromise between accuracy and efficiency, but the fixed resolution restricts the representation of small objects.
More recently, transformer-based architectures~\cite{zhao2021point,wu2022point,wu2024point,yang2023swin3d,yang2024swin3d++,lai2022stratified} have improved long-range dependency modeling, but their quadratic complexity constrains large-scale deployment.
Overall, the inherent sparsity and limited semantic abstraction of LiDAR data motivate the integration of complementary visual information to enhance contextual understanding.

\subsection{Multimodal 2D-3D Fusion for 3D Semantic Segmentation}


Fusing LiDAR and RGB improves 3D segmentation by coupling geometric precision with visual semantics~\cite{xu2021rpvnet,yan20222dpass,li2023mseg3d, li2019bidirectional,morerio2017minimal,vu2019advent}. Early dual-stream systems project points to the image plane and merge features via residual or distillation modules, whereas later approaches jointly optimize intra-modal learning and inter-modal interaction to enhance complementarity~\cite{an2024multimodality,an2025generalized,zhu2025rethinking}.  However, most existing methods implicitly assume that features from different modalities can be directly aligned without explicitly modeling their shared and modality-specific components. This often leads to suboptimal fusion, where complementary information is not fully utilized and modality-dependent noise may be introduced. Hierarchical and semantic-guided strategies refine structural coherence through adaptive weighting, and vision-language encoders provide contextual priors and open-vocabulary capability~\cite{liu2023uniseg,wu2024unidseg}. Despite these advances, many pipelines still depend on dense attention or direct concatenation, which raises computational cost and weakens interpretability. Performance degradation under illumination, weather, or sensor shifts further underscores the need for domain-adaptive multimodal fusion.

\subsection{Domain Adaptation and Generalization in 3D Semantic Segmentation}

Domain adaptation and generalization address distribution shifts in multimodal 3D perception by transferring from labeled sources to unlabeled or unseen targets. Representative methods promote cross-modal consistency via bidirectional learning or align class prototypes through adversarial objectives, while fusion-distillation and style-transfer strategies aim to improve target robustness~\cite{cao2024mopa, cardace2023exploiting, li2022cross,wu2023cross,jaritz2020xmuda,jaritz2022cross,liu2021adversarial,peng2021sparse,wu2025fusion,zhang2023mx2m, zhang2022self,wu2024unidseg,liang2025unidxmd}. Recent unified representations reduce modality and domain gaps with structured quantization and latent regularization. Nevertheless, many frameworks still treat fusion and adaptation separately, which limits semantic consistency across modalities and stability across domains; a unified formulation remains necessary.

\section{Methodology}

\subsection{Framework Overview}


The proposed framework is outlined in Fig.~\ref{fig:framework}. It adopts a dual-branch structure for 2D images and 3D point clouds, enabling reliable fusion and consistent generalization. The 3D branch uses sparse convolution and transformer layers~\cite{zhang2024sptnet} to extract geometric structure, while the 2D branch applies a SAM-based encoder~\cite{kirillov2023segment} to produce semantically coherent visual features. The resulting representations are projected into a shared space and decomposed into shared and private components. Shared features represent the semantic content that is consistent across modalities and are fused through the Shared Attention Fusion block. Private features retain the modality-dependent information needed to complement the shared representation. The fused output is decoded into detailed segmentation predictions, and the training objectives guide the network toward better alignment, clearer feature separation, and improved accuracy.

\subsection{Dual-Branch Feature Extraction}

To capture both fine-grained geometric structures and high-level semantic information, we design a dual-branch feature extraction architecture that jointly leverages 2D imagery and 3D point clouds.The two modalities are inherently complementary, as LiDAR provides precise spatial geometry whereas RGB imagery contributes dense contextual semantics. To achieve a robust multimodal representation, each branch is carefully designed to preserve its intrinsic advantages while maintaining consistent cross-modal alignment.

For the image branch, we adopt the Segment Anything Model (SAM) as the vision encoder due to its strong generalization ability and proven effectiveness in segmentation-oriented visual tasks. SAM has demonstrated exceptional capability in capturing object boundaries and semantic coherence across diverse datasets ~\cite{lee2025effective,wu2025every,wu2024unidseg}. By integrating the hierarchical visual tokens produced by the Segment Anything Model (SAM), the framework extracts multi-scale features that encode both local structural details and global semantic relationships. The resulting features are projected into a shared embedding space through a lightweight transformation layer, ensuring semantic consistency with the geometric features extracted from the 3D branch.

For the point cloud branch, we employ a Sparse Convolution-Transformer hybrid network (SPTNet), which efficiently captures fine-grained geometric variations through sparse convolutions and models long-range dependencies via transformer-based self-attention. This design allows the network to encode both local and contextual spatial relations, generating structurally complete and semantically aligned 3D representations. The features produced by both branches are normalized into a common latent space, forming the basis for subsequent shared-private feature decomposition and semantic fusion.

\subsection{Shared-Private Feature Decomposition and Semantic Fusion}

To structure the multimodal interaction and reduce redundancy during fusion, each modality is designed to be decomposed into a shared component that captures modality-invariant semantics and a private component that preserves modality-specific characteristics. This design is formalized by representing each encoded feature as the sum of its shared and private components,
\begin{equation}
\mathbf{F}^{m} = \mathbf{S}^{m} + \mathbf{R}^{m}, \quad m \in \{2\mathrm{D}, 3\mathrm{D}\},
\end{equation}
where $\mathbf{S}^{m}$ encodes semantic information shared across modalities, and $\mathbf{R}^{m}$ preserves modality-specific attributes rooted in either geometric structure or visual appearance. The shared and private components are obtained through learnable linear projections and normalization layers, providing a structured latent space for subsequent fusion.

To ensure that the decomposition is meaningfully achieved during training, two regularization constraints are introduced to guide the separation of shared and private subspaces. The first constraint encourages the shared components to encode consistent semantic structure across modalities by aligning their second-order statistics:
\begin{equation}
\label{eq2}
\mathcal{L}_{\mathrm{gram}} = \frac{1}{C_s^2}
\left\| \mathbf{S}^{2\mathrm{D}}_{512}{}^\top \mathbf{S}^{2\mathrm{D}}_{512}
- \mathbf{S}^{3\mathrm{D}}_{512}{}^\top \mathbf{S}^{3\mathrm{D}}_{512} \right\|_F^2.
\end{equation}
The second constraint promotes independence between private components by penalizing their cross-correlation,
\begin{equation}
\label{eq3}
\mathcal{L}_{\mathrm{diff}} = \frac{1}{C_p^2}
\left\| \mathbf{R}^{2\mathrm{D}}_{512}{}^\top \mathbf{R}^{3\mathrm{D}}_{512} \right\|_F^2,
\end{equation}
thereby reducing redundancy and preventing modality-specific variations from interfering with shared semantics.

With the decomposed representations obtained, the shared features $\mathbf{S}^{2\mathrm{D}}_{512}$ and $\mathbf{S}^{3\mathrm{D}}_{512}$ are integrated by a Shared Attention Fusion (SAF) module, which aggregates modality-invariant information according to their semantic relevance. The attention weights $\mathbf{A}$
are computed as
\begin{equation}
\mathbf{A}
= \mathrm{Softmax}\!\left( 
\frac{\mathbf{Q}(\mathbf{S}^{3\mathrm{D}}_{512})
\mathbf{K}(\mathbf{S}^{2\mathrm{D}}_{512})^{\top}}{\sqrt{d}}
\right),
\end{equation}
where $\mathbf{Q}(\cdot)$ and $\mathbf{K}(\cdot)$ denote linear transformations and $d$ is the feature dimension. The 3D features are assigned as queries to guide the point-level prediction process, whereas the 2D features serve as keys to deliver supplementary semantic information obtained from the image branch. The fused shared representation is then computed as
\begin{equation}
\label{eq5}
\mathbf{S}_{1024}
= \mathbf{A}\mathbf{V}(\mathbf{S}^{2\mathrm{D}}_{512})
+ (1 - \mathbf{A})\mathbf{V}(\mathbf{S}^{3\mathrm{D}}_{512}),
\end{equation}
where $\mathbf{V}(\cdot)$ denotes the value transformation. This aggregation balances the contribution of both modalities and maintains stable semantic correspondence in the shared space. For clarity, all feature dimensions are explicitly indicated in Eqs.~\eqref{eq2}-\eqref{eq5} and the SAF formulation.

\begin{figure}[htbp]
    \centering
    \vspace{-8pt}
    \includegraphics[width=0.6\linewidth]{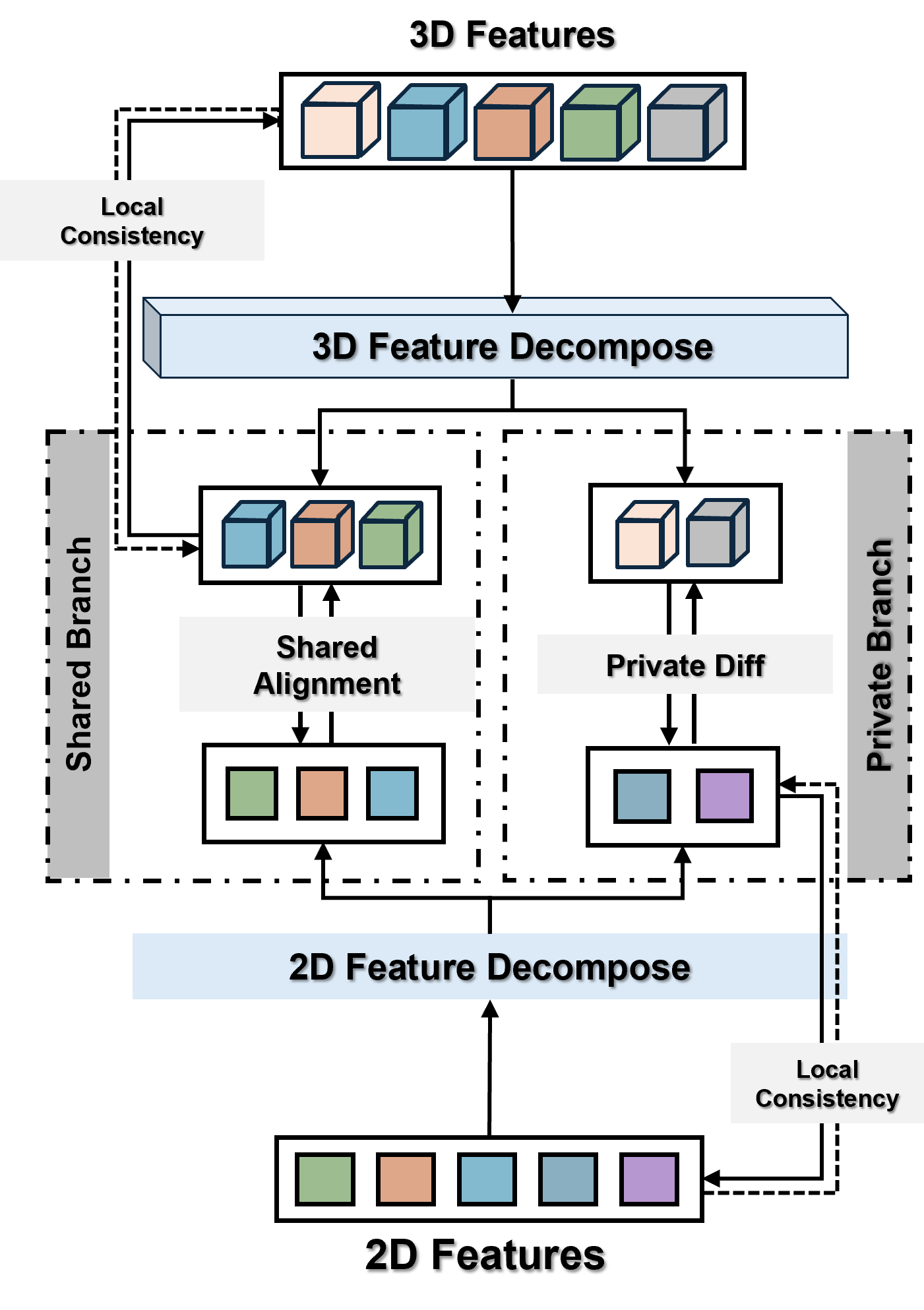}
    \caption{Architecture of the proposed shared-private feature decomposition.}
    \vspace{-8pt}
    \label{fig:Decompose}
\end{figure}

\begin{table*}[]

\centering
\caption{Quantitative results on the nuScenes validation set. Numbers for some baselines are taken from their original papers. }
\label{tab:nuscenes_results}
\resizebox{\linewidth}{!}{
\begin{tabular}{l c *{16}{c} c}
\toprule
\rowcolor{lightgray}
Method & Modality & barrier & bicycle & bus & car & construction & motorcycle & pedestrian & traffic cone & trailer & truck & driveable & other flat & sidewalk & terrain & manmade & vegetation & mIoU(\%) \\
\midrule
RangeNet++~\cite{milioto2019rangenet++}         & L  & 66.0 & 21.3 & 77.2 & 80.9 & 30.2 & 66.8 & 69.6 & 52.1 & 54.2 & 72.3 & 94.1 & 66.6 & 63.5 & 70.1 & 83.1 & 79.8 & 65.5 \\
PolarNet~\cite{zhang2020polarnet}           & L  & 74.7 & 28.2 & 85.3 & 90.9 & 35.1 & 77.5 & 71.3 & 58.8 & 57.4 & 76.1 & 96.5 & 71.1 & 74.7 & 74.0 & 87.3 & 85.7 & 71.0 \\
SalsaNext~\cite{cortinhal2020salsanext}          & L  & 74.8 & 34.1 & 85.9 & 88.4 & 42.2 & 72.4 & 72.2 & 63.1 & 61.3 & 76.5 & 96.0 & 70.8 & 71.2 & 71.5 & 86.7 & 84.4 & 72.2 \\
Cylinder3D~\cite{zhu2021cylindrical}         & L  & 76.4 & 40.3 & 91.3 & 93.8 & 51.3 & 78.0 & 78.9 & 64.9 & 62.1 & 84.4 & 96.8 & 71.6 & \underline{76.4} & 75.4 & \underline{90.5} & 87.4 & 76.1 \\
SFCNet~\cite{zheng2024spherical}             & L  & 76.7 & 40.4 & 89.5 & 91.3 & 46.7 & 82.0 & 78.1 & 65.8 & 69.4 & 80.6 & 96.6 & 71.6 & 74.5 & 74.9 & 89.0 & 87.5 & 75.9 \\
RangeVit~\cite{ando2023rangevit}           & L  & 75.5 & 40.7 & 88.3 & 90.1 & 49.3 & 79.3 & 77.2 & 66.3 & 65.2 & 80.0 & 96.4 & 71.4 & 73.8 & 73.8 & 89.9 & 87.2 & 75.2 \\
MemorySeg~\cite{li2023memoryseg}          & L  &   -  &   -  &   -  &   -  &   -  &   -  &   -  &   -  &   -  &   -  &   -  &   -  &   -  &   -  &   -  &   -  & 76.7 \\
\midrule
PMF~\cite{zhuang2021perception}                & LC & 74.1 & 46.6 & 89.8 & 92.1 & 57.0 & 77.7 & 80.9 & \textbf{70.9} & 64.6 & 82.9 & 95.5 & 73.3 & 73.6 & 74.8 & 89.4 & \underline{87.7} & 76.9 \\
2DPASS~\cite{yan20222dpass}             & LC & \underline{78.4} & 52.5 & 95.4 & 93.4 & 62.5 & 88.9 & \underline{83.1} & 68.1 & \underline{75.6} & \textbf{89.1} & \underline{96.9} & 75.5 & 76.3 & 75.7 & 89.1 & 86.9 & 80.5 \\
PTv3~\cite{wu2024point}               & L  &   -  &   -  &   -  &   -  &   -  &   -  &   -  &   -  &   -  &   -  &   -  &   -  &   -  &   -  &   -  &   -  & 80.2 \\
PTv3+PPT~\cite{wu2024point}           & L  &   -  &   -  &   -  &   -  &   -  &   -  &   -  &   -  &   -  &   -  &   -  &   -  &   -  &   -  &   -  &   -  & 81.2 \\

CSFNet (large)~\cite{luo2025csfnet} & LC & 77.8 & \underline{52.6 }& \textbf{96.6} & \underline{94.4} & \textbf{64.1} & \textbf{89.3} & 82.9 & 67.9 & \textbf{77.0} & \underline{87.7} & 96.8 & \underline{76.6} & 76.2 & \underline{76.1} & 89.3 & 86.7 & 80.8 \\
U2MKD~\cite{sun2024uni}           & L  &   -  &   -  &   -  &   -  &   -  &   -  &   -  &   -  &   -  &   -  &   -  &   -  &   -  &   -  &   -  &   -  & \textbf{83.1} \\
\midrule
\rowcolor{lightgray}
UniD-shift(ours) & LC & \textbf{78.9} & \textbf{52.7} & \underline{96.5} & 93.8 & \underline{63.6} & \underline{89.2} & \textbf{84.5} & \underline{68.6} & 72.9 & 87.6 & \textbf{97.1} & \textbf{77.9} & \textbf{77.8} & \textbf{77.7} & \textbf{91.3} & \textbf{88.9} & \underline{81.0}  \\

\bottomrule
\end{tabular}
}
\begin{minipage}{0.98\textwidth}\footnotesize
\textbf{Notes.} \textbf{L}: LiDAR-only. \textbf{LC}: LiDAR-Camera fusion. 
\end{minipage}
\end{table*}
Finally, the fused shared representation is concatenated with the 3D private features to produce the multimodal representation used for semantic prediction:
\begin{equation}
\mathbf{F}^{\mathrm{fused}} = [\mathbf{S}; \mathbf{R}^{3\mathrm{D}}],
\end{equation}
which is subsequently decoded into per-point semantic labels. By combining structured feature decomposition with attention-based fusion, the framework preserves modality-specific information while aligning semantic concepts across modalities, resulting in a compact and semantically coherent multimodal embedding.

\begin{table}[htbp]
\centering
\caption{Quantitative results on the nuScenes test set.}
\label{tab:nuScene_test}
\footnotesize
\setlength{\tabcolsep}{4pt}
\begin{tabular}{l c l c}
\toprule
\rowcolor{lightgray}
\multicolumn{2}{c}{LiDAR-only} & \multicolumn{2}{c}{LiDAR-Camera} \\
\cmidrule(r){1-2} \cmidrule(l){3-4}
Method & mIoU (\%) & Method & mIoU (\%) \\
\midrule
PolarNet~\cite{zhang2020polarnet}    & 69.4 & PMF~\cite{zhuang2021perception}        & 77.0 \\
Cylinder3D~\cite{zhou2020cylinder3d}  & 77.2 & 2D3DNet~\cite{genova2021learning}    & 80.0 \\
CMDFusion~\cite{cen2023cmdfusion}   & 80.8 & MSeg3D\cite{li2023mseg3d}     & 81.1 \\
LidarMultiNet~\cite{ye2023lidarmultinet}      & 81.4 & U2MKD\cite{sun2024uni}     & 84.2 \\
LiDARFormer~\cite{zhou2024lidarformer}  & 81.5 & TASeg~\cite{wu2024taseg}    & \textbf{84.6} \\
\rowcolor{lightgray} SphereFormer~\cite{lai2023spherical} & \textbf{81.9} & UniD-Shift (Ours) & 81.2 \\
\bottomrule
\vspace{-8pt}
\end{tabular}
\end{table}

\subsection{Training Objectives}

The training strategy is designed to jointly optimize semantic prediction accuracy, cross-modal consistency, and feature disentanglement. Accordingly, a set of complementary loss functions is introduced, each addressing a distinct learning objective. These components include semantic supervision for both modalities, cross-modal knowledge alignment, and regularization terms that stabilize the shared-private decomposition process. 

\noindent
\textbf{Shared-private regularization.}  
Two regularization terms are introduced to stabilize feature decomposition and maintain disentanglement between subspaces. The Gram alignment $\mathcal{L}{\mathrm{gram}}$ (Eq.~\ref{eq2}) enforces consistency between the shared representations of both modalities, ensuring that they capture coherent semantic structures. Meanwhile, the decorrelation $\mathcal{L}{\mathrm{diff}}$ (Eq.~\ref{eq3}) reduces redundancy across the private subspaces, promoting independence between modality-specific components.

\noindent
\textbf{Segmentation loss.}  
Semantic supervision is provided for both the 2D and 3D branches to ensure accurate category prediction within each modality. Each segmentation loss integrates a weighted cross-entropy term with a Lovász-Softmax loss, enhancing class balance and boundary precision:
\begin{equation}
\mathcal{L}_{\mathrm{seg}}^{m} = \mathcal{L}_{\mathrm{CE}}^{m} + \mathcal{L}_{\mathrm{Lovasz}}^{m}, \quad m \in \{2\mathrm{D}, 3\mathrm{D}\}.
\end{equation}
The 3D segmentation branch provides point-wise geometric supervision, while the 2D branch refines the fused prediction by leveraging dense pixel-level semantics.

\noindent
\textbf{Cross-modal knowledge distillation.}  
To maintain semantic consistency between modalities, a bidirectional Kullback-Leibler divergence is adopted to align the predictive distributions of the image and point cloud branches:
\begin{equation}
\mathcal{L}_{\mathrm{xm}} = \mathrm{KL}\!\left(\log p(\hat{\mathbf{y}}^{3\mathrm{D}}),\, p(\hat{\mathbf{y}}^{2\mathrm{D}})\right),
\end{equation}
where $\hat{\mathbf{y}}^{2\mathrm{D}}$ and $\hat{\mathbf{y}}^{3\mathrm{D}}$ denote the fused and point-level predictions, respectively. This objective promotes bidirectional semantic adaptation and mutual feature calibration.

\noindent
\textbf{Overall optimization.}  
All objectives are jointly optimized in an end-to-end manner. The total training loss is expressed as
\begin{equation}
\small
\mathcal{L}_{\mathrm{total}} = \mathcal{L}_{\mathrm{seg}}^{3\mathrm{D}} + \lambda_{\mathrm{seg2D}}\mathcal{L}_{\mathrm{seg}}^{2\mathrm{D}} + \lambda_{\mathrm{xm}}\mathcal{L}_{\mathrm{xm}} + \lambda_{\mathrm{gram}}\mathcal{L}_{\mathrm{gram}} + \lambda_{\mathrm{diff}}\mathcal{L}_{\mathrm{diff}},
\end{equation}
where the weighting coefficients $\lambda_{\mathrm{seg2D}}$, $\lambda_{\mathrm{xm}}$, $\lambda_{\mathrm{gram}}$, and $\lambda_{\mathrm{diff}}$ determine the relative contribution of each component. This joint optimization encourages balanced learning of semantic accuracy, modality alignment, and disentangled representation structure.

\section{Experiments and Analysis}

\subsection{Experimental Setup}

\textbf{Benchmarks.}  
Experiments are performed on the nuScenes~\cite{caesar2020nuscenes} and SemanticKITTI~\cite{behley2019semantickitti} benchmarks, both providing synchronized LiDAR-camera data with dense semantic labels. nuScenes offers multimodal driving sequences with 16 categories, while SemanticKITTI provides annotated LiDAR sequences with 19 categories. These datasets serve as the basis for evaluating the proposed framework in single-domain settings, enabling a comprehensive analysis of segmentation accuracy and multimodal fusion behavior.

\noindent\textbf{Model and Training Configuration.}  
All experiments are implemented in PyTorch and conducted on a distributed setup with eight NVIDIA RTX 4090 GPUs. The network is trained for 100 epochs using stochastic gradient descent with an initial learning rate of 0.24, momentum of 0.9, weight decay of $10^{-4}$, and a cosine annealing schedule. Random rotation, flipping, and uniform scaling are applied during data augmentation to improve generalization. The image branch adopts a SAM-based vision-language encoder, and the point cloud branch employs an SPTNet backbone. Both branches are jointly optimized through shared-private feature decomposition and semantic fusion to enhance cross-modal complementarity and maintain modality-specific distinctiveness.

\begin{figure}[htbp]
    \centering
    \includegraphics[width=1\linewidth]{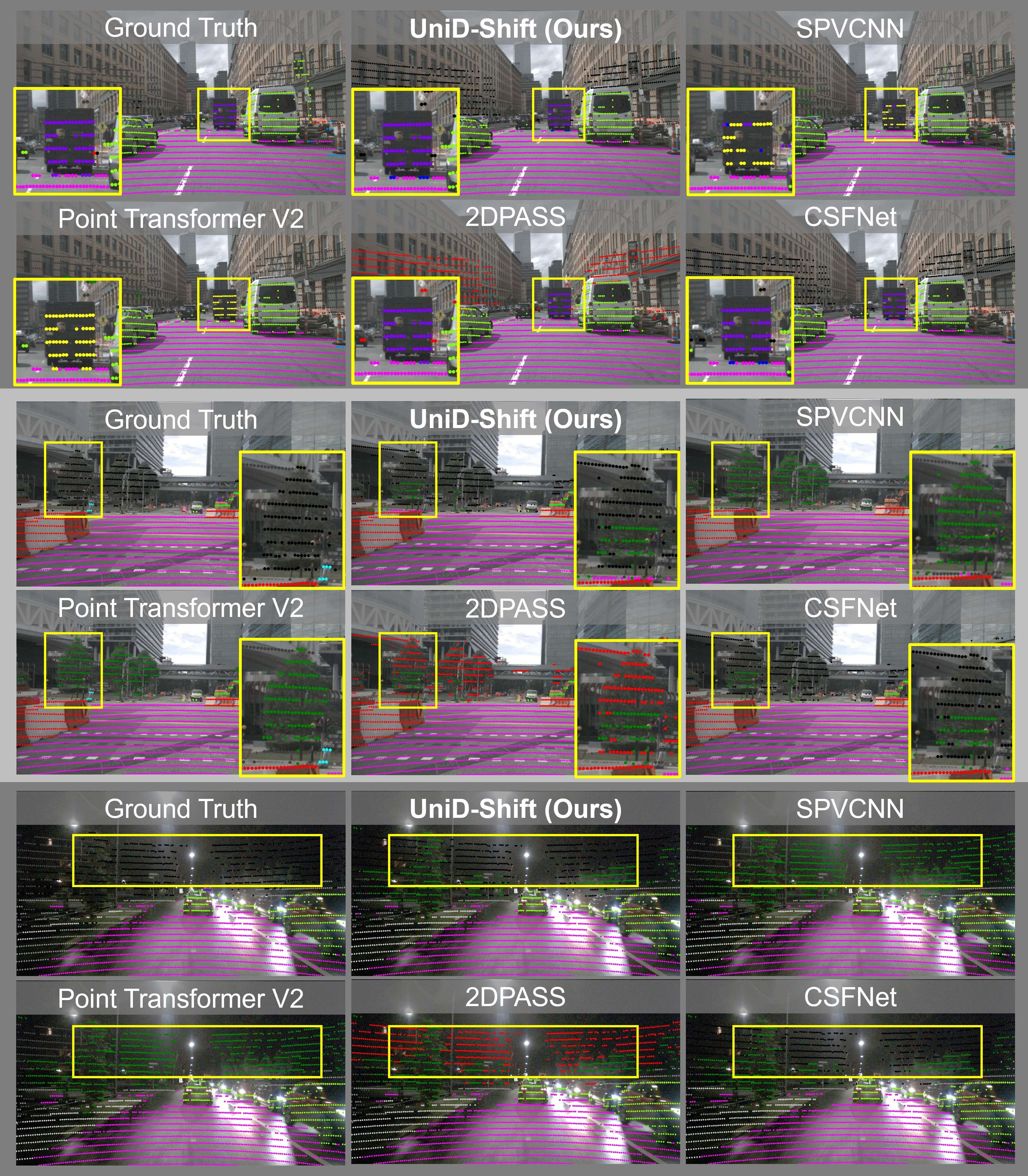}
    \caption{Visualization of object segmentation performance on the nuScenes validation set.}
    \vspace{-8pt}
    \label{fig:example}
\end{figure}

\subsection{Single-Domain Results}

We evaluate UniD-Shift under the single-domain setting on the nuScenes~\cite{caesar2020nuscenes} and SemanticKITTI~\cite{behley2019semantickitti} benchmarks. The results in Tables~\ref{tab:nuscenes_results}--\ref{tab:semkitti_val} indicate consistent improvements over existing multimodal segmentation frameworks. UniD-Shift achieves an mIoU of 81.0\% on nuScenes validation set (Table~\ref{tab:nuscenes_results}), 81.2\% on nuScenes test set (Tables~\ref{tab:nuScene_test},\ref{tab:nuscenes_test_results}), and 71.8\% on SemanticKITTI (Tables~\ref{tab:semkitti_val}), exceeding the performance of 2DPASS\cite{yan20222dpass}, CSFNet\cite{luo2025csfnet}, and other strong multimodal fusion baselines. The improvements extend across categories with different scales, densities, and surface characteristics, indicating that the fusion strategy preserves geometric detail while maintaining stable semantic structure. UniD-Shift integrates information from dense imagery and sparse LiDAR in a consistent manner, producing predictions that re main well organized rather than conflicting or overly smoothed. Results across both datasets further show that the model establishes a coherent multimodal representation that remains reliable in regions affected by occlusion, irregular sampling, or strong geometric variation.

\begin{figure}[h]
    \centering
    \includegraphics[width=\linewidth]{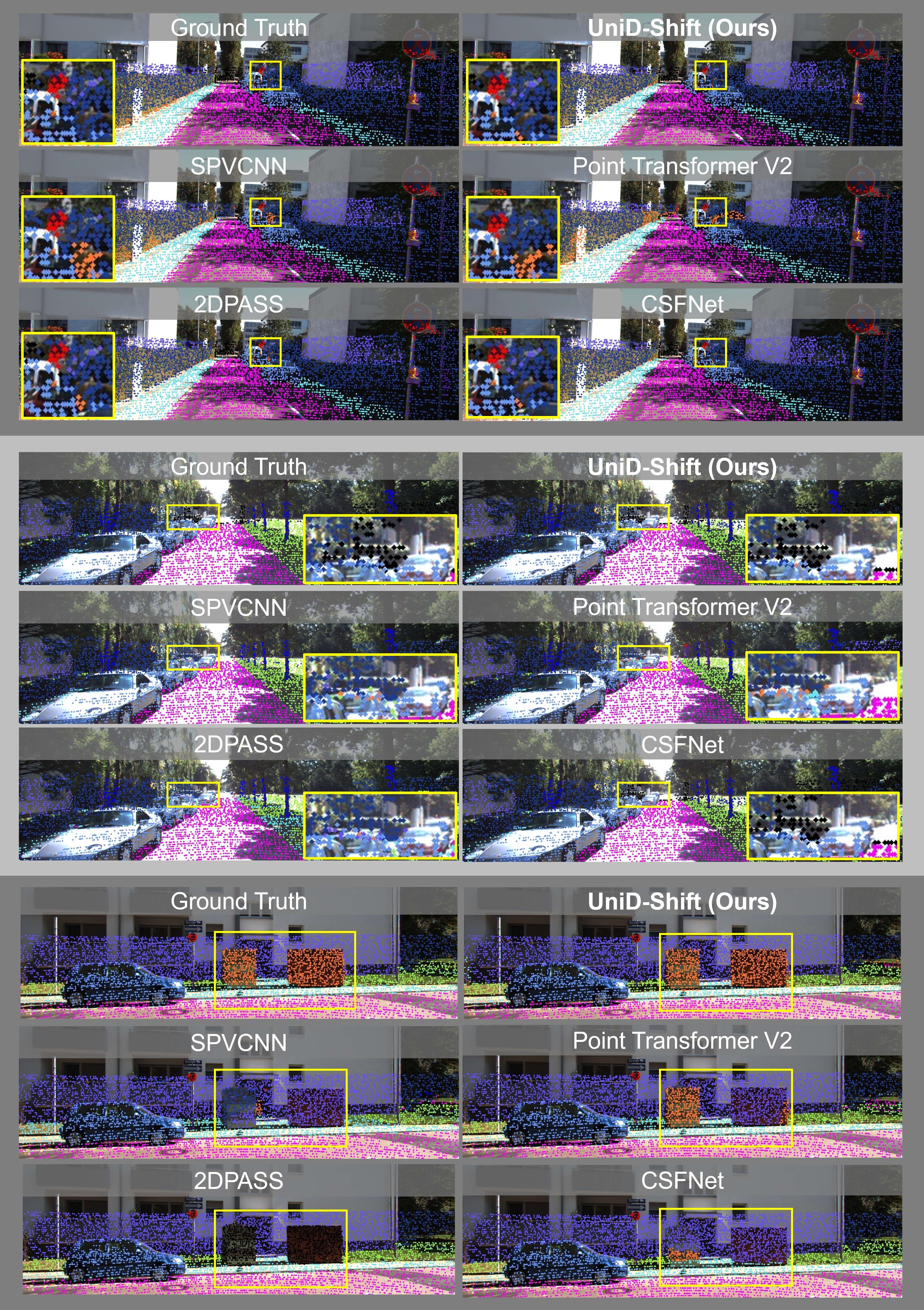}
    \caption{Visualization of object segmentation performance on the SemanticKitti validation set.}
    \label{fig:example2}
    \vspace{-8pt}
\end{figure}

The visual comparisons in Figs.~\ref{fig:example} and ~\ref{fig:example2} include both single-modality baselines (SPVCNN~\cite{tang2020searching}, Point Transformer V2~\cite{wu2022point}) and multimodal baselines (2DPASS~\cite{yan20222dpass}, CSFNet~\cite{luo2025csfnet}). The single-modality models retain strong geometric detail but often lose continuity when the point distribution becomes sparse. The multimodal baselines improve category coverage but still produce scattered or incomplete regions in scenes with complex depth changes or dense structural layouts.UniD-Shift delivers clearer object contours and more continuous semantic regions across both datasets. The predictions preserve structural integrity in areas where point density becomes irregular, while maintaining stable category transitions across adjacent regions. In particular, the fused representation reduces fragmented patches and isolated artifacts that frequently appear in baseline results. Moreover, the spatial layout remains coherent across the scene, even under strong depth variation or complex urban geometry. These observations indicate that UniD-Shift learns a resilient multimodal representation and maintains reliable performance in large-scale environments. The consistent advantage across diverse conditions suggests stable fusion across varying sensing conditions.

\begin{table*}[h]
\centering
\scriptsize
\setlength{\tabcolsep}{3pt}
\renewcommand{\arraystretch}{1.15}
\newcommand{\rot}[1]{\rotatebox{60}{\strut #1}}
\caption{Quantitative results on the SemanticKITTI validation set. 
Numbers for some baselines are taken from their original papers.}
\label{tab:semkitti_val}
\resizebox{\textwidth}{!}{%
\begin{tabular}{l c
*{19}{c}
c}
\toprule
\rowcolor{lightgray}
Method & Modality &
\rot{car} & \rot{bicycle} & \rot{motorcycle} & \rot{truck} & \rot{other vehicle} & \rot{person} &
\rot{bicyclist} & \rot{motorcyclist} & \rot{road} & \rot{parking} & \rot{sidewalk} & \rot{other ground} &
\rot{building} & \rot{fence} & \rot{vegetation} & \rot{trunk} & \rot{terrain} & \rot{pole} & \rot{traffic sign} &
mIoU (\%) \\
\midrule
RangeNet++~\cite{milioto2019rangenet++}             & L  & 89.4 & 26.5 & 48.4 & 33.9 & 26.7 & 54.8 & 69.4 & 0.0 & 92.9 & 37.0 & 69.9 & 0.0 & 83.4 & 51.0 & 83.3 & 54.0 & 68.1 & 49.8 & 34.0 & 51.2 \\
SqueezeSegV2~\cite{wu2019squeezesegv2}           & L  & 82.7 & 15.1 & 22.7 & 25.6 & 26.9 & 22.9 & 44.5 & 0.0 & 92.7 & 39.7 & 70.7 & 0.1 & 71.6 & 37.0 & 74.6 & 35.8 & 68.1 & 21.8 & 22.2 & 40.8 \\
SqueezeSegV3~\cite{xu2020squeezesegv3}           & L  & 87.1 & 34.3 & 48.6 & 47.5 & 47.1 & 58.1 & 53.8 & 0.0 & \underline{95.3} & 43.1 & 78.2 & 0.3 & 78.9 & 53.2 & 82.3 & 55.5 & 70.4 & 46.3 & 33.2 & 53.3 \\
RandLA-Net~\cite{hu2020randla}             & L  & 92.0 & 8.0 & 12.8 & 74.8 & 46.7 & 52.3 & 46.0 & 0.0 & 93.4 & 32.7 & 73.4 & 0.1 & 84.0 & 43.5 & 83.7 & 57.3 & 73.1 & 48.0 & 27.3 & 50.0 \\
SalsaNext~\cite{cortinhal2020salsanext}              & L  & 90.5 & 44.6 & 49.6 & 86.3 & 54.6 & 74.0 & 81.4 & 0.0 & 93.4 & 40.6 & 69.1 & 0.0 & 84.6 & 53.0 & 83.6 & 64.3 & 64.2 & 54.4 & 39.8 & 59.4 \\
MinkowskiNet~\cite{choy20194d}           & L  & 95.0 & 23.9 & 50.4 & 55.3 & 45.9 & 65.6 & 82.2 & 0.0 & 94.3& 43.7 & 76.4 & 0.0 & 87.9 & 57.4 & 87.4 & 67.7 & 71.5 & 63.5 & 43.6 & 58.5 \\
SPVCNN~\cite{tang2020searching}                 & L  & 96.5 & 44.8 & 63.1 & 59.9 & 64.3 & 12.0 & 86.0 & 0.0 & 93.9 & 42.4 & 75.9 & 0.0 & 88.8 & 59.1 & 88.0 & 67.5 & 73.0 & 63.5 & 44.3 & 62.3 \\
Cylinder3D~\cite{zhu2021cylindrical}             & L  & 96.4 & \underline{61.5} & 78.2 & 66.3 & 69.8 & \underline{80.8} & 93.3 & 0.0 & 94.9 & 41.5 & 78.0 & 1.4 & 87.5 & 50.0 & 86.7 & 72.2 & 68.8 & 63.0 & 42.1 & 64.9 \\
FA-ResNet~\cite{zhan2023fa}              & L  & 93.7 & 30.6 & 33.2 & 33.9 & 27.2 & 51.6 & 30.6 & \textbf{18.7} & 91.3 & \textbf{65.2} & 75.5 & \textbf{26.0} & 90.6 & 62.7 & 83.1 & 63.5 & 66.3 & 51.0 & 48.4 & 55.7 \\
PTv3~\cite{wu2024point}                   & L  & - & - & - & - & - & - & - & - & - & - & - & - & - & - & - & - & - & - & - & 70.8 \\
PTv3+PPT~\cite{wu2024point}               & L  & - & - & - & - & - & - & - & - & - & - & - & - & - & - & - & - & - & - & - & \textbf{72.3} \\
\midrule
PMF~\cite{zhuang2021perception}                    & LC & 95.4 & 47.8 & 62.9 & 68.4 & 75.2 & 78.9 & 71.6 & 0.0 & \textbf{96.4} & 43.5 & 80.5 & 0.1 & 88.7 & 60.1 & \underline{88.6} & \underline{72.7} & 75.3 & 65.5 & 43.0 & 63.9 \\
2DPASS~\cite{yan20222dpass}                 & LC & \underline{97.2} & 55.3 & 78.3 & 90.9 & 70.7 & 80.7 & \underline{93.5} & 0.1 & 94.3 & 53.8 & \underline{82.0} & 12.1 & 93.0 & \underline{71.8} & \textbf{89.1} & 72.3 & \underline{75.7} & 66.0 & \underline{54.2} & 70.0 \\
MSeg3D~\cite{li2023mseg3d}                 & LC & - & - & - & - & - & - & - & - & - & - & - & - & - & - & - & - & - & - & - & 69.0 \\
U2MKD~\cite{sun2024uni}& LC & - & - & - & - & - & - & - & - & - & - & - & - & - & - & - & - & - & - & - & 69.6 \\
UniSeg~\cite{liu2023uniseg}                 & LC & - & - & - & - & - & - & - & - & - & - & - & - & - & - & - & - & - & - & - & 71.3 \\
CSFNet (large)~\cite{luo2025csfnet}  & LC & 
\textbf{97.6} & \textbf{61.5} & \underline{88.5} & \textbf{96.5} & \textbf{82.8} & \textbf{82.4} & \textbf{93.6} & 2.2 & 94.8 & 52.1 & \textbf{83.0} & 0.0 & \textbf{93.4} & \textbf{73.7} & \textbf{89.1} & \textbf{72.8} & \underline{75.6} & \textbf{66.3} & 52.3 & 71.5 \\

\midrule
\rowcolor{lightgray}
UniD-shift(ours) & LC & 
 96.8 & 59.4 & \textbf{89.2} & \underline{92.3} & \underline{80.1} & \textbf{82.4} & \textbf{93.6} & \underline{13.1} & 94.7 & \underline{61.5} & 78.3 & \underline{14.2} & \underline{93.3} & 60.2 & 83.2 & 71.7 & \textbf{76.0} & \underline{66.1} & \textbf{57.8} & \underline{71.8} \\
\bottomrule
\end{tabular}}
\vspace{-0.5em}
\begin{minipage}{0.98\textwidth}\footnotesize
\textbf{Notes.} \textbf{L}: LiDAR-only. \textbf{LC}: LiDAR-Camera fusion. 
\end{minipage}
\end{table*}



\subsection{Ablation Study}


As shown in Table~\ref{tab:AS}, we perform an ablation study on the nuScenes to investigate how each design choice influences the overall performance. Replacing SPVCNN with SPTNet results in a clear improvement in geometric feature quality, which reflects the benefit of transformer-based modeling in capturing long-range structural dependencies. The substitution of the 2D encoder with SAM brings an additional performance gain, as the foundation model supplies richer semantic signals together with more stable pixel-level representations. Although these upgrades enhance the representational capacity of each modality, their effect remains limited when the interaction between branches relies on KL-based alignment. This observation underscores the need for a dedicated fusion mechanism that can fully exploit the complementary nature of 2D and 3D information.

Introducing the shared-private fusion module produces a more substantial improvement, since this component reorganizes multimodal features by separating modality-invariant semantics from modality-specific information.
This separation reduces interference between branches and yields a more coherent correspondence across modalities.
The configuration that integrates SAM, SPTNet, and the shared-private fusion achieves the best performance in the ablation study, reaching an mIoU of \textbf{74.8\%} in Table~\ref{tab:AS}. The comparison indicates that the performance improvement is primarily attributable to the structured fusion mechanism. This design regulates the interaction between heterogeneous modalities and strengthens the integration of geometric information with semantic content. For more ablation details, see Supplementary Section~\ref{Additional Ablation Studies}, Table~\ref{tab:FAS}.

\subsection{Cross-Domain Extension and Generalization}

We conducted a representative cross-domain adaptation experiment on the nuScenes dataset under the \textbf{USA $\rightarrow$ Singapore} setting to examine the robustness of UniD-Shift under clear changes in urban layout and sensing conditions.  The experiment follows a standard cross-domain protocol in which the network is optimized with annotated data from the USA domain and evaluated on the Singapore domain. Both domains share the same sensing configuration and pipeline, ensuring that variations arise from environmental differences rather than data preparation.





\begin{table}[h]
\centering
\caption{Comprehensive ablation study on backbone, fusion strategy, SAF query direction, and loss contributions.More detailed ablation studies are provided in Supplementary Section~\ref{Additional Ablation Studies} (Table~\ref{tab:FAS}).}
\label{tab:AS}

\setlength{\tabcolsep}{6pt}
\scriptsize
\resizebox{\linewidth}{!}{%
\begin{tabular}{lcccc}
\toprule
\rowcolor{lightgray}
Exp & 2D Module & 3D Module & Fusion / Setting & mIoU (\%) \\
\midrule
\multicolumn{5}{l}{\textit{Backbone + Fusion Ablation}} \\
\midrule
1 (Baseline) & ResNet & SPVCNN & KL & 68.2\textcolor{red}{($\downarrow$6.6)} \\
2            & ResNet & SPTNet & KL & 70.1\textcolor{red}{($\downarrow$4.7)} \\
3            & SAM    & SPTNet & KL & 73.4\textcolor{red}{($\downarrow$1.4)} \\
4 (Ours)     & SAM    & SPTNet & Shared-Private Fusion & \textbf{74.8} \\
\midrule
\multicolumn{5}{l}{\textit{SAF Query Direction Ablation}} \\
\midrule
5            & SAM & SPTNet & 2D$\rightarrow$3D & 70.9\textcolor{red}{($\downarrow$3.9)} \\
6            & SAM & SPTNet & 2D$\leftrightarrow$3D & 71.5\textcolor{red}{($\downarrow$3.3)} \\
7 (Ours)     & SAM & SPTNet & 3D$\rightarrow$2D & \textbf{74.8} \\
\midrule
\multicolumn{5}{l}{\textit{Loss Ablation}} \\
\midrule
8            & SAM & SPTNet & w/o Gram & 73.7\textcolor{red}{($\downarrow$1.1)} \\
9            & SAM & SPTNet & w/o Decorr & 74.4\textcolor{red}{($\downarrow$0.4)} \\
10           & SAM & SPTNet & w/o Gram+Decorr & 73.4\textcolor{red}{($\downarrow$1.4)} \\
\bottomrule
\end{tabular}}
\end{table}

As shown in Table~\ref{tab:nuscenes_usasing}, UniD-Shift reaches an mIoU of \textbf{74.5\%} on the USA to Singapore benchmark and outperforms xMUDA (69.4\%), UniDSeg (72.9\%), and UniDxMD (74.3\%). The advantage appears consistently across both 2D and 3D branches, suggesting that the shared-private formulation improves cross-modal alignment stability under domain shifts. The margin over fusion-based adaptation models indicates robust representations under variations in layout, objects, and background.

The overall trend in Table~\ref{tab:nuscenes_usasing} shows that UniD-Shift delivers higher scores not only in the combined multimodal prediction but also in the unimodal components, revealing that the fusion strategy strengthens each branch instead of amplifying domain bias. The results point to a representation that absorbs complementary information from the two modalities in a balanced manner and maintains semantic consistency under geographic and structural variation. This behavior suggests that UniD-Shift forms a more transferable feature space, enabling reliable segmentation performance when applied to unseen urban environments.

\begin{table}[]
\setlength\tabcolsep{10pt}
    \centering 
    \renewcommand\arraystretch{1}

\caption{Cross-domain results on \textbf{nuScenes (USA/Sing.)}. 
“Uni.” denotes uni-modal UDA, and “Cross.” denotes cross-modal UDA. Comparison highlights the effectiveness of multimodal adaptation.
}
\label{tab:nuscenes_usasing}
  \scriptsize
  \resizebox{\linewidth}{!}{%
\begin{tabular}{llrrr}
\toprule
\rowcolor{lightgray}
\textbf{Mod.} & \textbf{Method} & \textbf{2D} & \textbf{3D} & \textbf{xM} \\
\midrule
\multirow{3}{*}{\textbf{Uni.}} 
& MinEnt~\cite{vu2019advent}        & 57.6 & 61.5 & 66.0 \\
& logCORAL~\cite{morerio2017minimal}    & 64.4 & 63.2 & 69.4 \\
& PL~\cite{li2019bidirectional}                & 62.0 & 64.8 & 70.4 \\
\midrule
\multirow{10}{*}{\textbf{Cross.}} 
& xMUDA~\cite{jaritz2022cross}          & 64.4 & 63.2 & 69.4 \\
& AUDA~\cite{liu2021adversarial}           & 63.2 & 61.9 & 67.5 \\
& DsCML~\cite{peng2021sparse}         & 62.7 & 61.8 & 67.7 \\
& Dual-Cross~\cite{li2022cross}& 60.9 & 61.2 & 68.4 \\
& SSE-xMUDA~\cite{zhang2022self}  & 66.7 & 65.4 & 70.8 \\
& BFtD~\cite{wu2023cross}            & 63.7 & 62.2 & 69.4 \\
& FtD++~\cite{wu2025fusion}          & 69.7 & 64.6 & 69.8 \\
& UniDSeg~\cite{wu2024unidseg} & 67.2 & 67.6 & 72.9\\
& MM2D3D~\cite{cardace2023exploiting}        & 71.7 & 66.8 & 72.4 \\
& UniDxMD~\cite{liang2025unidxmd}                    & \underline{73.2} & \underline{68.5} & \underline{74.3} \\
\midrule
\rowcolor{lightgray}
 & \textbf{UniD-Shift (Ours)} & \textbf{73.3} & \textbf{68.8} & \textbf{74.5} \\
\bottomrule
\end{tabular}}
\end{table}

\section{Discussion}

\subsection{Interpretability and Fusion Efficiency}

We further evaluate the balance between accuracy and efficiency to examine the practical value of the framework. The shared-private decomposition organizes multimodal features into distinct semantic roles and reduces interference during fusion, leading to a stable optimization process. Table~\ref{tab:semkitti_eff} shows that UniD-Shift reaches an mIoU of \textbf{71.8\%} on SemanticKITTI with a latency of \textbf{240 ms}. The results indicate that the frozen SAM encoder increases the overall parameter count while keeping the trainable portion compact, and the inference time remains lower than other SAM-based designs. This pattern reflects a fusion mechanism that limits computation and restricts cross-modal exchange to essential information. The evidence suggests UniD-Shift achieves a balanced trade-off between accuracy and computational cost, yielding an efficient and stable multimodal solution.

\begin{table}[htbp]
\centering
\caption{Comparisons between efficiency (run-time) and accuracy on the SemanticKITTI validation set. A clear trade-off between latency and segmentation performance can be observed.}
\label{tab:semkitti_eff}

\setlength{\tabcolsep}{6pt}
\scriptsize
\resizebox{\linewidth}{!}{%
\begin{tabular}{l r r r}
\toprule
\rowcolor{lightgray}
Method & \#Params (M) & Latency (ms) & mIoU (\%) $\uparrow$ \\
\midrule
MinkowskiNet~\cite{choy20194d}      & 21.7            & 48  & 61.1 \\
Cylinder3D~\cite{zhu2021cylindrical}        & 56.3            & 75  & 65.9 \\
SPVCNN~\cite{tang2020searching}            & 21.8            & 52  & 63.8 \\
UniSeg~\cite{liu2023uniseg}            & 147.6           & 145 & 71.3 \\
CSFNet~\cite{luo2025csfnet}   & 45.4            & 122 & 71.5 \\
UniDSeg (CLIP)~\cite{wu2024unidseg}    & 350.96 (38.69)  & 214 & N/A \\
UniDSeg (SAM)~\cite{wu2024unidseg}     & 353.44 (38.69)  & 296 & N/A \\
\midrule
\rowcolor{lightgray}
UniD-Shift (SAM)  & 359.80 (35.20)  & 240 &  71.8\\\bottomrule
\end{tabular}}

\end{table}

\subsection{Cross-Domain Robustness and Generalization}

The proposed framework exhibits strong robustness in cross-domain evaluation. In the USA$\rightarrow$Singapore adaptation setting of the nuScenes dataset, UniD-Shift attains the highest performance among all compared methods, reaching an mIoU of \textbf{74.5\%} as shown in Table~\ref{tab:nuscenes_usasing}. The improvement reflects the ability of the shared-private decomposition to separate stable semantic information from domain-dependent variations, thereby reducing the influence of differences in spatial layout, road structure, and environmental appearance across cities. The model forms a unified latent representation that supports reliable transfer from labeled source data to the unlabeled target domain and preserves semantic consistency throughout the prediction process. The overall trend indicates that UniD-Shift sustains high segmentation accuracy while maintaining stable behavior across  urban environments, demonstrating its effectiveness for large-scale multimodal domain adaptation.

\subsection{Limitations}

Despite its strong performance, the framework depends on reliable geometric correspondence between LiDAR and camera data, which may constrain its use in scenes where the two modalities are not precisely synchronized. In addition, employing a large visual backbone increases memory demand during training and reduces scalability on devices with limited computational resources. Future research will investigate more compact fusion structures and adaptive alignment mechanisms to lower overhead and enhance suitability for real-time and large-scale applications.

\section{Conclusion}

In this paper, we propose UniD-Shift, a unified multimodal framework for 3D semantic segmentation that integrates image and point cloud information through shared-private feature decomposition. The method integrates SAM-driven visual priors with SPTNet-based geometric encoding to form a coherent latent space capable of capturing both semantic organization and spatial structure. Extensive experiments on nuScenes and SemanticKITTI demonstrate that UniD-Shift delivers superior segmentation accuracy while maintaining computational efficiency and training stability throughout the entire process. Cross-domain evaluations further confirm that the learned representations effectively under diverse environmental and sensor conditions, establishing the framework as a reliable and interpretable solution for the large-scale multimodal 3D scene understanding.

\section*{Acknowledgments}
This work is supported by the Young Scientists Fund of the National Natural Science Foundation of China  (Grant No. 42401567), Guangdong Provincial Project	(Grant No. 2024QN11G095), AI Research and Learning Base of Urban Culture, Guangdong Provincial Department of Education	(Grant No. 2023WZJD008), and the Open Fund of the Technology Innovation Center for 3D Real Scene Construction and Urban Refined Governance, Ministry of Natural Resources (Grant No. 2024PF-1)


{
\small
\bibliographystyle{ieeenat_fullname}
\bibliography{main}

@String(ECCV= {Eur. Conf. Comput. Vis.})

@String(AAAI = {AAAI})

@String(ECCV  = {ECCV})

@inproceedings{chen2023svqnet,
  title={Svqnet: Sparse voxel-adjacent query network for 4d spatio-temporal lidar semantic segmentation},
  author={Chen, Xuechao and Xu, Shuangjie and Zou, Xiaoyi and Cao, Tongyi and Yeung, Dit-Yan and Fang, Lu},
  booktitle={Proceedings of the IEEE/CVF International Conference on Computer Vision},
  pages={8569--8578},
  year={2023}
}

@inproceedings{wu2024taseg,
  title={Taseg: Temporal aggregation network for lidar semantic segmentation},
  author={Wu, Xiaopei and Hou, Yuenan and Huang, Xiaoshui and Lin, Binbin and He, Tong and Zhu, Xinge and Ma, Yuexin and Wu, Boxi and Liu, Haifeng and Cai, Deng and others},
  booktitle={Proceedings of the IEEE/CVF Conference on Computer Vision and Pattern Recognition},
  pages={15311--15320},
  year={2024}
}

@article{sun2024uni,
  title={Uni-to-multi modal knowledge distillation for bidirectional lidar-camera semantic segmentation},
  author={Sun, Tianfang and Zhang, Zhizhong and Tan, Xin and Peng, Yong and Qu, Yanyun and Xie, Yuan},
  journal={IEEE Transactions on Pattern Analysis and Machine Intelligence},
  volume={46},
  number={12},
  pages={11059--11072},
  year={2024},
  publisher={IEEE}
}

@inproceedings{li2023mseg3d,
  title={Mseg3d: Multi-modal 3d semantic segmentation for autonomous driving},
  author={Li, Jiale and Dai, Hang and Han, Hao and Ding, Yong},
  booktitle={Proceedings of the IEEE/CVF conference on computer vision and pattern recognition},
  pages={21694--21704},
  year={2023}
}

@inproceedings{genova2021learning,
  title={Learning 3d semantic segmentation with only 2d image supervision},
  author={Genova, Kyle and Yin, Xiaoqi and Kundu, Abhijit and Pantofaru, Caroline and Cole, Forrester and Sud, Avneesh and Brewington, Brian and Shucker, Brian and Funkhouser, Thomas},
  booktitle={2021 International Conference on 3D Vision (3DV)},
  pages={361--372},
  year={2021},
  organization={IEEE}
}

@inproceedings{zhuang2021perception,
  title={Perception-aware multi-sensor fusion for 3d lidar semantic segmentation},
  author={Zhuang, Zhuangwei and Li, Rong and Jia, Kui and Wang, Qicheng and Li, Yuanqing and Tan, Mingkui},
  booktitle={Proceedings of the IEEE/CVF international conference on computer vision},
  pages={16280--16290},
  year={2021}
}

@inproceedings{lai2023spherical,
  title={Spherical transformer for lidar-based 3d recognition},
  author={Lai, Xin and Chen, Yukang and Lu, Fanbin and Liu, Jianhui and Jia, Jiaya},
  booktitle={Proceedings of the IEEE/CVF conference on computer vision and pattern recognition},
  pages={17545--17555},
  year={2023}
}

@inproceedings{ye2023lidarmultinet,
  title={Lidarmultinet: Towards a unified multi-task network for lidar perception},
  author={Ye, Dongqiangzi and Zhou, Zixiang and Chen, Weijia and Xie, Yufei and Wang, Yu and Wang, Panqu and Foroosh, Hassan},
  booktitle={Proceedings of the AAAI Conference on Artificial Intelligence},
  volume={37},
  number={3},
  pages={3231--3240},
  year={2023}
}

@inproceedings{zhou2024lidarformer,
  title={Lidarformer: A unified transformer-based multi-task network for lidar perception},
  author={Zhou, Zixiang and Ye, Dongqiangzi and Chen, Weijia and Xie, Yufei and Wang, Yu and Wang, Panqu and Foroosh, Hassan},
  booktitle={2024 IEEE International Conference on Robotics and Automation (ICRA)},
  pages={14740--14747},
  year={2024},
  organization={IEEE}
}

@inproceedings{tang2020searching,
  title={Searching efficient 3d architectures with sparse point-voxel convolution},
  author={Tang, Haotian and Liu, Zhijian and Zhao, Shengyu and Lin, Yujun and Lin, Ji and Wang, Hanrui and Han, Song},
  booktitle={European conference on computer vision},
  pages={685--702},
  year={2020},
  organization={Springer}
}

@article{cen2023cmdfusion,
  title={Cmdfusion: Bidirectional fusion network with cross-modality knowledge distillation for lidar semantic segmentation},
  author={Cen, Jun and Zhang, Shiwei and Pei, Yixuan and Li, Kun and Zheng, Hang and Luo, Maochun and Zhang, Yingya and Chen, Qifeng},
  journal={IEEE Robotics and Automation Letters},
  volume={9},
  number={1},
  pages={771--778},
  year={2023},
  publisher={IEEE}
}

@inproceedings{yan20222dpass,
  title={2dpass: 2d priors assisted semantic segmentation on lidar point clouds},
  author={Yan, Xu and Gao, Jiantao and Zheng, Chaoda and Zheng, Chao and Zhang, Ruimao and Cui, Shuguang and Li, Zhen},
  booktitle={European conference on computer vision},
  pages={677--695},
  year={2022},
  organization={Springer}
}

@article{zhou2020cylinder3d,
  title={Cylinder3d: An effective 3d framework for driving-scene lidar semantic segmentation},
  author={Zhou, Hui and Zhu, Xinge and Song, Xiao and Ma, Yuexin and Wang, Zhe and Li, Hongsheng and Lin, Dahua},
  journal={arXiv preprint arXiv:2008.01550},
  year={2020}
}

@inproceedings{zhang2020polarnet,
  title={Polarnet: An improved grid representation for online lidar point clouds semantic segmentation},
  author={Zhang, Yang and Zhou, Zixiang and David, Philip and Yue, Xiangyu and Xi, Zerong and Gong, Boqing and Foroosh, Hassan},
  booktitle={Proceedings of the IEEE/CVF conference on computer vision and pattern recognition},
  pages={9601--9610},
  year={2020}
}

@inproceedings{milioto2019rangenet++,
  title={Rangenet++: Fast and accurate lidar semantic segmentation},
  author={Milioto, Andres and Vizzo, Ignacio and Behley, Jens and Stachniss, Cyrill},
  booktitle={2019 IEEE/RSJ international conference on intelligent robots and systems (IROS)},
  pages={4213--4220},
  year={2019},
  organization={IEEE}
}

@inproceedings{cortinhal2020salsanext,
  title={Salsanext: Fast, uncertainty-aware semantic segmentation of lidar point clouds},
  author={Cortinhal, Tiago and Tzelepis, George and Erdal Aksoy, Eren},
  booktitle={International Symposium on Visual Computing},
  pages={207--222},
  year={2020},
  organization={Springer}
}

@inproceedings{zhu2021cylindrical,
  title={Cylindrical and asymmetrical 3d convolution networks for lidar segmentation},
  author={Zhu, Xinge and Zhou, Hui and Wang, Tai and Hong, Fangzhou and Ma, Yuexin and Li, Wei and Li, Hongsheng and Lin, Dahua},
  booktitle={Proceedings of the IEEE/CVF conference on computer vision and pattern recognition},
  pages={9939--9948},
  year={2021}
}

@inproceedings{zhu2025rethinking,
  title={Rethinking end-to-end 2d to 3d scene segmentation in gaussian splatting},
  author={Zhu, Runsong and Qiu, Shi and Liu, Zhengzhe and Hui, Ka-Hei and Wu, Qianyi and Heng, Pheng-Ann and Fu, Chi-Wing},
  booktitle={Proceedings of the IEEE/CVF Conference on Computer Vision and Pattern Recognition},
  pages={3656--3665},
  year={2025}
}

@inproceedings{an2025generalized,
  title={Generalized few-shot 3d point cloud segmentation with vision-language model},
  author={An, Zhaochong and Sun, Guolei and Liu, Yun and Li, Runjia and Han, Junlin and Konukoglu, Ender and Belongie, Serge},
  booktitle={Proceedings of the IEEE/CVF Conference on Computer Vision and Pattern Recognition},
  pages={16997--17007},
  year={2025}
}

@article{zheng2024spherical,
  title={Spherical frustum sparse convolution network for lidar point cloud semantic segmentation},
  author={Zheng, Yu and Wang, Guangming and Liu, Jiuming and Pollefeys, Marc and Wang, Hesheng},
  journal={Advances in Neural Information Processing Systems},
  volume={37},
  pages={121827--121858},
  year={2024}
}

@inproceedings{ando2023rangevit,
  title={Rangevit: Towards vision transformers for 3d semantic segmentation in autonomous driving},
  author={Ando, Angelika and Gidaris, Spyros and Bursuc, Andrei and Puy, Gilles and Boulch, Alexandre and Marlet, Renaud},
  booktitle={Proceedings of the IEEE/CVF conference on computer vision and pattern recognition},
  pages={5240--5250},
  year={2023}
}

@inproceedings{li2023memoryseg,
  title={Memoryseg: Online lidar semantic segmentation with a latent memory},
  author={Li, Enxu and Casas, Sergio and Urtasun, Raquel},
  booktitle={Proceedings of the IEEE/CVF International Conference on Computer Vision},
  pages={745--754},
  year={2023}
}

@inproceedings{wu2024point,
  title={Point transformer v3: Simpler faster stronger},
  author={Wu, Xiaoyang and Jiang, Li and Wang, Peng-Shuai and Liu, Zhijian and Liu, Xihui and Qiao, Yu and Ouyang, Wanli and He, Tong and Zhao, Hengshuang},
  booktitle={Proceedings of the IEEE/CVF conference on computer vision and pattern recognition},
  pages={4840--4851},
  year={2024}
}

@article{luo2025csfnet,
  title={CSFNet: Cross-modal Semantic Focus Network for Sematic Segmentation of Large-Scale Point Clouds},
  author={Luo, Yang and Han, Ting and Liu, Yujun and Su, Jinhe and Chen, Yiping and Li, Jinyuan and Wu, Yundong and Cai, Guorong},
  journal={IEEE Transactions on Geoscience and Remote Sensing},
  year={2025},
  publisher={IEEE}
}

@inproceedings{wu2019squeezesegv2,
  title={Squeezesegv2: Improved model structure and unsupervised domain adaptation for road-object segmentation from a lidar point cloud},
  author={Wu, Bichen and Zhou, Xuanyu and Zhao, Sicheng and Yue, Xiangyu and Keutzer, Kurt},
  booktitle={2019 international conference on robotics and automation (ICRA)},
  pages={4376--4382},
  year={2019},
  organization={IEEE}
}

@inproceedings{xu2020squeezesegv3,
  title={Squeezesegv3: Spatially-adaptive convolution for efficient point-cloud segmentation},
  author={Xu, Chenfeng and Wu, Bichen and Wang, Zining and Zhan, Wei and Vajda, Peter and Keutzer, Kurt and Tomizuka, Masayoshi},
  booktitle={European Conference on Computer Vision},
  pages={1--19},
  year={2020},
  organization={Springer}
}

@inproceedings{hu2020randla,
  title={Randla-net: Efficient semantic segmentation of large-scale point clouds},
  author={Hu, Qingyong and Yang, Bo and Xie, Linhai and Rosa, Stefano and Guo, Yulan and Wang, Zhihua and Trigoni, Niki and Markham, Andrew},
  booktitle={Proceedings of the IEEE/CVF conference on computer vision and pattern recognition},
  pages={11108--11117},
  year={2020}
}

@inproceedings{choy20194d,
  title={4d spatio-temporal convnets: Minkowski convolutional neural networks},
  author={Choy, Christopher and Gwak, JunYoung and Savarese, Silvio},
  booktitle={Proceedings of the IEEE/CVF conference on computer vision and pattern recognition},
  pages={3075--3084},
  year={2019}
}

@article{zhan2023fa,
  title={FA-ResNet: Feature affine residual network for large-scale point cloud segmentation},
  author={Zhan, Lixin and Li, Wei and Min, Weidong},
  journal={International Journal of Applied Earth Observation and Geoinformation},
  volume={118},
  pages={103259},
  year={2023},
  publisher={Elsevier}
}

@inproceedings{liu2023uniseg,
  title={Uniseg: A unified multi-modal lidar segmentation network and the openpcseg codebase},
  author={Liu, Youquan and Chen, Runnan and Li, Xin and Kong, Lingdong and Yang, Yuchen and Xia, Zhaoyang and Bai, Yeqi and Zhu, Xinge and Ma, Yuexin and Li, Yikang and others},
  booktitle={Proceedings of the IEEE/CVF International Conference on Computer Vision},
  pages={21662--21673},
  year={2023}
}

@article{wu2024unidseg,
  title={Unidseg: Unified cross-domain 3d semantic segmentation via visual foundation models prior},
  author={Wu, Yao and Xing, Mingwei and Zhang, Yachao and Luo, Xiaotong and Xie, Yuan and Qu, Yanyun},
  journal={Advances in Neural Information Processing Systems},
  volume={37},
  pages={101223--101249},
  year={2024}
}

@inproceedings{caesar2020nuscenes,
  title={nuscenes: A multimodal dataset for autonomous driving},
  author={Caesar, Holger and Bankiti, Varun and Lang, Alex H and Vora, Sourabh and Liong, Venice Erin and Xu, Qiang and Krishnan, Anush and Pan, Yu and Baldan, Giancarlo and Beijbom, Oscar},
  booktitle={Proceedings of the IEEE/CVF conference on computer vision and pattern recognition},
  pages={11621--11631},
  year={2020}
}

@inproceedings{behley2019semantickitti,
  title={Semantickitti: A dataset for semantic scene understanding of lidar sequences},
  author={Behley, Jens and Garbade, Martin and Milioto, Andres and Quenzel, Jan and Behnke, Sven and Stachniss, Cyrill and Gall, Jurgen},
  booktitle={Proceedings of the IEEE/CVF international conference on computer vision},
  pages={9297--9307},
  year={2019}
}

@inproceedings{vu2019advent,
  title={Advent: Adversarial entropy minimization for domain adaptation in semantic segmentation},
  author={Vu, Tuan-Hung and Jain, Himalaya and Bucher, Maxime and Cord, Matthieu and P{\'e}rez, Patrick},
  booktitle={Proceedings of the IEEE/CVF conference on computer vision and pattern recognition},
  pages={2517--2526},
  year={2019}
}

@article{morerio2017minimal,
  title={Minimal-entropy correlation alignment for unsupervised deep domain adaptation},
  author={Morerio, Pietro and Cavazza, Jacopo and Murino, Vittorio},
  journal={arXiv preprint arXiv:1711.10288},
  year={2017}
}

@inproceedings{li2019bidirectional,
  title={Bidirectional learning for domain adaptation of semantic segmentation},
  author={Li, Yunsheng and Yuan, Lu and Vasconcelos, Nuno},
  booktitle={Proceedings of the IEEE/CVF conference on computer vision and pattern recognition},
  pages={6936--6945},
  year={2019}
}

@article{jaritz2022cross,
  title={Cross-modal learning for domain adaptation in 3d semantic segmentation},
  author={Jaritz, Maximilian and Vu, Tuan-Hung and De Charette, Raoul and Wirbel, {\'E}milie and P{\'e}rez, Patrick},
  journal={IEEE Transactions on Pattern Analysis and Machine Intelligence},
  volume={45},
  number={2},
  pages={1533--1544},
  year={2022},
  publisher={IEEE}
}

@article{liu2021adversarial,
  title={Adversarial unsupervised domain adaptation for 3D semantic segmentation with multi-modal learning},
  author={Liu, Wei and Luo, Zhiming and Cai, Yuanzheng and Yu, Ying and Ke, Yang and Junior, Jos{\'e} Marcato and Gon{\c{c}}alves, Wesley Nunes and Li, Jonathan},
  journal={ISPRS Journal of Photogrammetry and Remote Sensing},
  volume={176},
  pages={211--221},
  year={2021},
  publisher={Elsevier}
}

@inproceedings{peng2021sparse,
  title={Sparse-to-dense feature matching: Intra and inter domain cross-modal learning in domain adaptation for 3d semantic segmentation},
  author={Peng, Duo and Lei, Yinjie and Li, Wen and Zhang, Pingping and Guo, Yulan},
  booktitle={Proceedings of the IEEE/CVF International Conference on Computer Vision},
  pages={7108--7117},
  year={2021}
}

@inproceedings{li2022cross,
  title={Cross-domain and cross-modal knowledge distillation in domain adaptation for 3d semantic segmentation},
  author={Li, Miaoyu and Zhang, Yachao and Xie, Yuan and Gao, Zuodong and Li, Cuihua and Zhang, Zhizhong and Qu, Yanyun},
  booktitle={Proceedings of the 30th ACM International Conference on Multimedia},
  pages={3829--3837},
  year={2022}
}

@inproceedings{zhang2022self,
  title={Self-supervised exclusive learning for 3d segmentation with cross-modal unsupervised domain adaptation},
  author={Zhang, Yachao and Li, Miaoyu and Xie, Yuan and Li, Cuihua and Wang, Cong and Zhang, Zhizhong and Qu, Yanyun},
  booktitle={Proceedings of the 30th ACM International Conference on Multimedia},
  pages={3338--3346},
  year={2022}
}

@inproceedings{wu2023cross,
  title={Cross-modal unsupervised domain adaptation for 3d semantic segmentation via bidirectional fusion-then-distillation},
  author={Wu, Yao and Xing, Mingwei and Zhang, Yachao and Xie, Yuan and Fan, Jianping and Shi, Zhongchao and Qu, Yanyun},
  booktitle={Proceedings of the 31st ACM International Conference on Multimedia},
  pages={490--498},
  year={2023}
}

@article{wu2025fusion,
  title={Fusion-then-Distillation: Toward Cross-modal Positive Distillation for Domain Adaptive 3D Semantic Segmentation},
  author={Wu, Yao and Xing, Mingwei and Zhang, Yachao and Xie, Yuan and Peng, Kaibei and Qu, Yanyun},
  journal={IEEE Transactions on Circuits and Systems for Video Technology},
  year={2025},
  publisher={IEEE}
}

@inproceedings{cardace2023exploiting,
  title={Exploiting the complementarity of 2d and 3d networks to address domain-shift in 3d semantic segmentation},
  author={Cardace, Adriano and Ramirez, Pierluigi Zama and Salti, Samuele and Di Stefano, Luigi},
  booktitle={Proceedings of the IEEE/CVF Conference on Computer Vision and Pattern Recognition},
  pages={98--109},
  year={2023}
}

@inproceedings{liang2025unidxmd,
  title={UniDxMD: Towards Unified Representation for Cross-Modal Unsupervised Domain Adaptation in 3D Semantic Segmentation},
  author={Liang, Zhengyin and Yin, Hui and Liang, Min and Du, Qianqian and Yang, Ying and Huang, Hua},
  booktitle={Proceedings of the IEEE/CVF International Conference on Computer Vision},
  pages={20346--20356},
  year={2025}
}

@article{zhang2024sptnet,
  title={SPTNet: Sparse convolution and transformer network for woody and foliage components separation from point clouds},
  author={Zhang, Shuai and Chen, Yiping and Wang, Biao and Pan, Dong and Zhang, Wuming and Li, Aiguang},
  journal={IEEE Transactions on Geoscience and Remote Sensing},
  volume={62},
  pages={1--18},
  year={2024},
  publisher={IEEE}
}

@inproceedings{kirillov2023segment,
  title={Segment anything},
  author={Kirillov, Alexander and Mintun, Eric and Ravi, Nikhila and Mao, Hanzi and Rolland, Chloe and Gustafson, Laura and Xiao, Tete and Whitehead, Spencer and Berg, Alexander C and Lo, Wan-Yen and others},
  booktitle={Proceedings of the IEEE/CVF international conference on computer vision},
  pages={4015--4026},
  year={2023}
}

@inproceedings{qu2025end,
  title={An end-to-end robust point cloud semantic segmentation network with single-step conditional diffusion models},
  author={Qu, Wentao and Wang, Jing and Gong, YongShun and Huang, Xiaoshui and Xiao, Liang},
  booktitle={Proceedings of the Computer Vision and Pattern Recognition Conference},
  pages={27325--27335},
  year={2025}
}

@article{zhang2024point,
  title={Point and voxel cross perception with lightweight cosformer for large-scale point cloud semantic segmentation},
  author={Zhang, Shuai and Wang, Biao and Chen, Yiping and Zhang, Shuhang and Zhang, Wuming},
  journal={International Journal of Applied Earth Observation and Geoinformation},
  volume={131},
  pages={103951},
  year={2024},
  publisher={Elsevier}
}

@inproceedings{zhao2025bfanet,
  title={BFANet: Revisiting 3D Semantic Segmentation with Boundary Feature Analysis},
  author={Zhao, Weiguang and Zhang, Rui and Wang, Qiufeng and Cheng, Guangliang and Huang, Kaizhu},
  booktitle={Proceedings of the Computer Vision and Pattern Recognition Conference},
  pages={29395--29405},
  year={2025}
}

@article{guo2020deep,
  title={Deep learning for 3d point clouds: A survey},
  author={Guo, Yulan and Wang, Hanyun and Hu, Qingyong and Liu, Hao and Liu, Li and Bennamoun, Mohammed},
  journal={IEEE transactions on pattern analysis and machine intelligence},
  volume={43},
  number={12},
  pages={4338--4364},
  year={2020},
  publisher={IEEE}
}

@inproceedings{kolodiazhnyi2024oneformer3d,
  title={Oneformer3d: One transformer for unified point cloud segmentation},
  author={Kolodiazhnyi, Maxim and Vorontsova, Anna and Konushin, Anton and Rukhovich, Danila},
  booktitle={Proceedings of the IEEE/CVF Conference on Computer Vision and Pattern Recognition},
  pages={20943--20953},
  year={2024}
}

@article{luo2025paseg,
  title={PASeg: positional-guided segmenter with multimodal semantic alignment for enhancing urban scene 3D semantic segmentation},
  author={Luo, Yang and Han, Ting and Zhang, Xiaorong and Liu, Yujun and Zhu, Duxin and Li, Jinyuan and Chen, Yiping and Wu, Yundong and Cai, Guorong and Piao, Yingchao and others},
  journal={International Journal of Digital Earth},
  volume={18},
  number={1},
  pages={2528811},
  year={2025},
  publisher={Taylor \& Francis}
}

@inproceedings{xu2021rpvnet,
  title={Rpvnet: A deep and efficient range-point-voxel fusion network for lidar point cloud segmentation},
  author={Xu, Jianyun and Zhang, Ruixiang and Dou, Jian and Zhu, Yushi and Sun, Jie and Pu, Shiliang},
  booktitle={Proceedings of the IEEE/CVF international conference on computer vision},
  pages={16024--16033},
  year={2021}
}

@article{an2024multimodality,
  title={Multimodality helps few-shot 3d point cloud semantic segmentation},
  author={An, Zhaochong and Sun, Guolei and Liu, Yun and Li, Runjia and Wu, Min and Cheng, Ming-Ming and Konukoglu, Ender and Belongie, Serge},
  journal={arXiv preprint arXiv:2410.22489},
  year={2024}
}

@article{deng2021multi,
  title={From multi-view to hollow-3D: Hallucinated hollow-3D R-CNN for 3D object detection},
  author={Deng, Jiajun and Zhou, Wengang and Zhang, Yanyong and Li, Houqiang},
  journal={IEEE Transactions on Circuits and Systems for Video Technology},
  volume={31},
  number={12},
  pages={4722--4734},
  year={2021},
  publisher={IEEE}
}

@article{hamdimvtn,
  title={MVTN: Multi-View Transformation Network for 3D Shape Recognition Supplementary Material},
  author={Hamdi, Abdullah and Giancola, Silvio and Ghanem, Bernard}
}

@inproceedings{liang2018deep,
  title={Deep continuous fusion for multi-sensor 3d object detection},
  author={Liang, Ming and Yang, Bin and Wang, Shenlong and Urtasun, Raquel},
  booktitle={Proceedings of the European conference on computer vision (ECCV)},
  pages={641--656},
  year={2018}
}

@inproceedings{su2015multi,
  title={Multi-view convolutional neural networks for 3d shape recognition},
  author={Su, Hang and Maji, Subhransu and Kalogerakis, Evangelos and Learned-Miller, Erik},
  booktitle={Proceedings of the IEEE international conference on computer vision},
  pages={945--953},
  year={2015}
}

@inproceedings{yang2018pixor,
  title={Pixor: Real-time 3d object detection from point clouds},
  author={Yang, Bin and Luo, Wenjie and Urtasun, Raquel},
  booktitle={Proceedings of the IEEE conference on Computer Vision and Pattern Recognition},
  pages={7652--7660},
  year={2018}
}

@inproceedings{graham20183d,
  title={3d semantic segmentation with submanifold sparse convolutional networks},
  author={Graham, Benjamin and Engelcke, Martin and Van Der Maaten, Laurens},
  booktitle={Proceedings of the IEEE conference on computer vision and pattern recognition},
  pages={9224--9232},
  year={2018}
}

@inproceedings{hou20193d,
  title={3d-sis: 3d semantic instance segmentation of rgb-d scans},
  author={Hou, Ji and Dai, Angela and Nie{\ss}ner, Matthias},
  booktitle={Proceedings of the IEEE/CVF conference on computer vision and pattern recognition},
  pages={4421--4430},
  year={2019}
}

@inproceedings{jiang2020pointgroup,
  title={Pointgroup: Dual-set point grouping for 3d instance segmentation},
  author={Jiang, Li and Zhao, Hengshuang and Shi, Shaoshuai and Liu, Shu and Fu, Chi-Wing and Jia, Jiaya},
  booktitle={Proceedings of the IEEE/CVF conference on computer vision and Pattern recognition},
  pages={4867--4876},
  year={2020}
}

@inproceedings{lang2019pointpillars,
  title={Pointpillars: Fast encoders for object detection from point clouds},
  author={Lang, Alex H and Vora, Sourabh and Caesar, Holger and Zhou, Lubing and Yang, Jiong and Beijbom, Oscar},
  booktitle={Proceedings of the IEEE/CVF conference on computer vision and pattern recognition},
  pages={12697--12705},
  year={2019}
}

@inproceedings{maturana2015voxnet,
  title={Voxnet: A 3d convolutional neural network for real-time object recognition},
  author={Maturana, Daniel and Scherer, Sebastian},
  booktitle={2015 IEEE/RSJ international conference on intelligent robots and systems (IROS)},
  pages={922--928},
  year={2015},
  organization={IEEE}
}

@inproceedings{song2016deep,
  title={Deep sliding shapes for amodal 3d object detection in rgb-d images},
  author={Song, Shuran and Xiao, Jianxiong},
  booktitle={Proceedings of the IEEE conference on computer vision and pattern recognition},
  pages={808--816},
  year={2016}
}

@article{yan2018second,
  title={Second: Sparsely embedded convolutional detection},
  author={Yan, Yan and Mao, Yuxing and Li, Bo},
  journal={Sensors},
  volume={18},
  number={10},
  pages={3337},
  year={2018},
  publisher={Multidisciplinary Digital Publishing Institute}
}

@inproceedings{qi2017pointnet,
  title={Pointnet: Deep learning on point sets for 3d classification and segmentation},
  author={Qi, Charles R and Su, Hao and Mo, Kaichun and Guibas, Leonidas J},
  booktitle={Proceedings of the IEEE conference on computer vision and pattern recognition},
  pages={652--660},
  year={2017}
}

@article{qi2017pointnet++,
  title={Pointnet++: Deep hierarchical feature learning on point sets in a metric space},
  author={Qi, Charles Ruizhongtai and Yi, Li and Su, Hao and Guibas, Leonidas J},
  journal={Advances in neural information processing systems},
  volume={30},
  year={2017}
}

@inproceedings{thomas2019kpconv,
  title={Kpconv: Flexible and deformable convolution for point clouds},
  author={Thomas, Hugues and Qi, Charles R and Deschaud, Jean-Emmanuel and Marcotegui, Beatriz and Goulette, Fran{\c{c}}ois and Guibas, Leonidas J},
  booktitle={Proceedings of the IEEE/CVF international conference on computer vision},
  pages={6411--6420},
  year={2019}
}

@inproceedings{zhao2021point,
  title={Point transformer},
  author={Zhao, Hengshuang and Jiang, Li and Jia, Jiaya and Torr, Philip HS and Koltun, Vladlen},
  booktitle={Proceedings of the IEEE/CVF international conference on computer vision},
  pages={16259--16268},
  year={2021}
}

@article{wu2022point,
  title={Point transformer v2: Grouped vector attention and partition-based pooling},
  author={Wu, Xiaoyang and Lao, Yixing and Jiang, Li and Liu, Xihui and Zhao, Hengshuang},
  journal={Advances in Neural Information Processing Systems},
  volume={35},
  pages={33330--33342},
  year={2022}
}

@article{yang2023swin3d,
  title={Swin3d: A pretrained transformer backbone for 3d indoor scene understanding},
  author={Yang, Yu-Qi and Guo, Yu-Xiao and Xiong, Jian-Yu and Liu, Yang and Pan, Hao and Wang, Peng-Shuai and Tong, Xin and Guo, Baining},
  journal={arXiv preprint arXiv:2304.06906},
  year={2023}
}

@article{yang2024swin3d++,
  title={Swin3D++: Effective Multi-Source Pretraining for 3D Indoor Scene Understanding},
  author={Yang, Yu-Qi and Guo, Yu-Xiao and Liu, Yang},
  journal={arXiv preprint arXiv:2402.14215},
  year={2024}
}

@inproceedings{lai2022stratified,
  title={Stratified transformer for 3d point cloud segmentation},
  author={Lai, Xin and Liu, Jianhui and Jiang, Li and Wang, Liwei and Zhao, Hengshuang and Liu, Shu and Qi, Xiaojuan and Jia, Jiaya},
  booktitle={Proceedings of the IEEE/CVF conference on computer vision and pattern recognition},
  pages={8500--8509},
  year={2022}
}

@inproceedings{zhou2021panoptic,
  title={Panoptic-polarnet: Proposal-free lidar point cloud panoptic segmentation},
  author={Zhou, Zixiang and Zhang, Yang and Foroosh, Hassan},
  booktitle={Proceedings of the IEEE/CVF conference on computer vision and pattern recognition},
  pages={13194--13203},
  year={2021}
}

@inproceedings{cao2024mopa,
  title={Mopa: Multi-modal prior aided domain adaptation for 3d semantic segmentation},
  author={Cao, Haozhi and Xu, Yuecong and Yang, Jianfei and Yin, Pengyu and Yuan, Shenghai and Xie, Lihua},
  booktitle={2024 IEEE International Conference on Robotics and Automation (ICRA)},
  pages={9463--9470},
  year={2024},
  organization={IEEE}
}

@inproceedings{jaritz2020xmuda,
  title={xmuda: Cross-modal unsupervised domain adaptation for 3d semantic segmentation},
  author={Jaritz, Maximilian and Vu, Tuan-Hung and Charette, Raoul de and Wirbel, Emilie and P{\'e}rez, Patrick},
  booktitle={Proceedings of the IEEE/CVF conference on computer vision and pattern recognition},
  pages={12605--12614},
  year={2020}
}

@inproceedings{zhang2023mx2m,
  title={Mx2m: masked cross-modality modeling in domain adaptation for 3d semantic segmentation},
  author={Zhang, Boxiang and Wang, Zunran and Ling, Yonggen and Guan, Yuanyuan and Zhang, Shenghao and Li, Wenhui},
  booktitle={Proceedings of the AAAI Conference on Artificial Intelligence},
  volume={37},
  number={3},
  pages={3401--3409},
  year={2023}
}

@inproceedings{lee2025effective,
  title={Effective SAM Combination for Open-Vocabulary Semantic Segmentation},
  author={Lee, Minhyeok and Cho, Suhwan and Lee, Jungho and Yang, Sunghun and Choi, Heeseung and Kim, Ig-Jae and Lee, Sangyoun},
  booktitle={Proceedings of the Computer Vision and Pattern Recognition Conference},
  pages={26081--26090},
  year={2025}
}

@inproceedings{wu2025every,
  title={Every SAM Drop Counts: Embracing Semantic Priors for Multi-Modality Image Fusion and Beyond},
  author={Wu, Guanyao and Liu, Haoyu and Fu, Hongming and Peng, Yichuan and Liu, Jinyuan and Fan, Xin and Liu, Risheng},
  booktitle={Proceedings of the Computer Vision and Pattern Recognition Conference},
  pages={17882--17891},
  year={2025}
}
}

\clearpage
\appendix
\renewcommand{\thesection}{\Alph{section}}
\renewcommand{\thesubsection}{\thesection\arabic{subsection}}

\setcounter{page}{1}
\maketitlesupplementary

\section{Additional Ablation Studies}
\label{Additional Ablation Studies}

To examine the contribution of each architectural choice, we conduct a comprehensive set of ablation experiments summarised in Table~\ref{tab:FAS}. The results reveal several consistent patterns. Replacing ResNet with SAM leads to a substantial improvement, with the gain reaching more than five percentage points when paired with SPVCNN. This trend reflects the benefit of dense semantic structure extracted by the foundation model. The transition from SPVCNN to SPTNet also produces a measurable increase, indicating that long-range geometric aggregation enhances the stability of point-wise representation. When both SAM and SPTNet are used simultaneously, the unimodal baseline becomes considerably stronger and forms a reliable basis for evaluating the contribution of fusion.

The second group of experiments focuses on the influence of the shared-private fusion design. When applied to weaker unimodal features, the fusion module alters the accuracy only slightly, yet its effect becomes more pronounced once the backbone features attain a higher level of semantic and geometric completeness. The comparison between experiments that differ only in the fusion choice demonstrates this trend clearly: the shift from KL-based alignment to shared-private fusion consistently produces improvement on top of both SAM-based and SPTNet-based settings. This behaviour indicates that the fusion mechanism benefits from well-structured modality inputs and reorganises them into a more coherent representation.

\begin{table}[h]
\centering
\caption{Full ablation studies evaluating the influence of the 2D encoder, 3D backbone, and fusion strategy on segmentation accuracy. }
\label{tab:FAS}

\setlength{\tabcolsep}{6pt}
\scriptsize
\resizebox{\linewidth}{!}{%
\begin{tabular}{lcccc}
\toprule
\rowcolor{lightgray}
Exp & 2D Module & 3D Module & Fusion & mIoU (\%) \\
\midrule
1(Baseline)     & ResNet     & SPVCNN     & KL  &  68.2\\
2     & ResNet     & SPTNet     & KL  &  70.1\\
3     & SAM        & SPVCNN     & KL  &  73.0\\
4     & SAM        & SPTNet     & KL  &  73.4\\
5     & ResNet     & SPVCNN     & Shared-Private fusion  &  Not Converged(64.8)\\
6     & ResNet     & SPTNet     & Shared-Private fusion  &  70.3\\
7     & SAM        & SPVCNN     & Shared-Private fusion  &  74.1\\
\rowcolor{lightgray}
8 (Ours) & SAM     & SPTNet     & Shared-Private fusion &  \textbf{74.8}\\
\bottomrule
\end{tabular}
}
\end{table}

The final configuration attains the highest accuracy at 74.8\%. The margin over the strongest unimodal setting confirms that the improvement is not solely attributable to the use of SAM or SPTNet. Instead, the interaction created by the shared-private formulation provides additional refinement, enabling the network to maintain stable semantic relations across modalities. The combined results illustrate how encoder strength and fusion structure reinforce one another and highlight the importance of coordinated multimodal design.

\section{Network Architecture Details}
\label{Network Architecture Details}

\subsection{Overall Architecture Overview}

The proposed framework consists of a dual-branch encoder, a shared-private feature decomposition module, a lightweight fusion block, and a 3D segmentation head. The 2D branch processes the RGB image through a frozen vision foundation model, while the 3D branch extracts geometric structure using a sparse convolution-transformer backbone. Both branches are projected into a unified latent space and decomposed into shared and private components. The fusion block aggregates the shared features to form a consistent multimodal representation, which is decoded to produce point-level predictions.

\subsection{2D Branch Details}

The 2D branch adopts the SAM ViT-L vision backbone, using the official pretrained weights sam\_vit\_l\_0b3195.pth. The encoder is kept frozen during training to retain stable visual priors and reduce optimization complexity. The model extracts hierarchical image features and produces dense representations that serve as the visual input to the multimodal pipeline.

Since SAM outputs high-dimensional visual tokens, a projection head is applied to map the features into the unified latent space. A linear layer followed by layer normalization compresses the SAM output into a 256-dimensional feature vector for each pixel aligned with valid LiDAR projections. This representation ensures that the 2D branch provides semantically rich information while maintaining compatibility with the 3D geometric features. The projected features are then used as the input to the shared-private decomposition module.

\subsection{3D branch details}

The 3D branch is built upon the SPTNet encoder, which couples sparse convolution with transformer modeling to capture geometric structure at multiple spatial scales. Sparse convolution operates on voxelized point clouds while preserving the irregular distribution of LiDAR points. Only non-empty voxels participate in computation, reducing the cost of 3D processing and maintaining the spatial layout of the scene. The encoder contains six SpConv blocks, each responsible for aggregating local geometry within a limited neighborhood. These blocks strengthen the description of fine structural patterns, yet their locality restricts the ability to capture long-range relations common in large outdoor environments.

To address this limitation, each SpConv block is followed by a transformer block that extends the receptive field beyond the convolutional neighborhood. The feature map from sparse convolution is linearly projected to form query, key, and value tensors, and multi-head attention is applied to model interactions across distant regions. This mechanism enhances the representation of the global structure and supports the integration of contexts that span multiple object scales. The transformer output is combined with the corresponding convolutional feature map through a skip connection, producing a unified representation that maintains both local detail and global coherence. By stacking six pairs of SpConv and transformer blocks, the encoder generates a hierarchical geometric feature pyramid, which serves as the 3D input to the shared-private fusion module described in the main paper.


\section{Implementation Details}
\label{sec:implementation_details}

\subsection{Training Configuration}

All experiments are conducted on eight NVIDIA RTX~4090 GPUs with 24~GB memory. The model adopts a hidden dimension of 256 and six hierarchical scales $\{2,4,8,16,16,16\}$ as specified in the configuration file. The 2D branch employs a frozen SAM encoder without any parameter updates, while the 3D branch uses an SPTNet backbone trained end to end.  During training, the loader applies geometric augmentation to point clouds and photometric perturbations to images, while the projection indices between 2D pixels and 3D points are precomputed to ensure stable multimodal alignment.

All trainable parameters belong to the 3D backbone and the fusion module, while the SAM encoder remains frozen. The total training loss follows the formulation in the main paper and includes 3D segmentation, 2D segmentation, cross-modal KL alignment, and the two decomposition terms. The loss weights are set as $\lambda_{\text{seg2D}}=1$, $\lambda_{\text{xm}}=1$, $\lambda_{\text{gram}}=0.05$, and $\lambda_{\text{diff}}=0.05$. Optimization uses stochastic gradient descent with a learning rate of $0.24$, momentum of $0.9$, and a weight decay of $10^{-4}$, together with cosine annealing over 80 epochs. Mixed-precision training is enabled, and each GPU handles a batch of 8 samples. This configuration ensures stable convergence for all experiments reported in the main paper.

\subsection{Inference and Runtime Measurement}

Inference is performed without test-time augmentation. Sparse convolutions reconstruct point-level features using the stored inverse voxel mapping, and semantic predictions are obtained by applying an $\arg\max$ operation on the class logits. The runtime reported in the main paper is measured using the same hardware environment and identical batch configuration to ensure fair comparison. All measurements reflect end-to-end forward time, including feature extraction, fusion, and final decoding.

\section{Detailed Performance Evaluation}
Table~\ref{tab:nuscenes_test_results} presents a detailed per-class evaluation on the nuScenes test set. Compared with LiDAR-only methods, multimodal approaches (LC) generally achieve stronger performance, highlighting the benefit of incorporating image cues for semantic understanding.  UniD-Shift achieves competitive overall performance, with clear improvements over several multimodal baselines in categories that require cross-modal reasoning, such as bus, motorcycle, and pedestrian. In addition, the method maintains stable predictions across both geometry-dominant classes (e.g., barrier and terrain) and appearance-sensitive categories (e.g., traffic cone and vegetation), demonstrating balanced semantic modeling.  These results further validate that the proposed shared-private decomposition effectively integrates complementary information from 2D and 3D modalities, leading to robust performance across diverse semantic classes.

\label{Detailed Performance Evaluation}
\begin{table*}[]

\centering
\caption{Quantitative results on the nuScenes test set. Numbers for some baselines are taken from their original papers. }
\label{tab:nuscenes_test_results}
\resizebox{\linewidth}{!}{
\begin{tabular}{l c *{16}{c} c}
\toprule
\rowcolor{lightgray}
Method & Modality & barrier & bicycle & bus & car & construction & motorcycle & pedestrian & traffic cone & trailer & truck & driveable & other flat & sidewalk & terrain & manmade & vegetation & mIoU(\%) \\
\midrule
PolarNet~\cite{zhang2020polarnet}& L & 69.4 & 72.2 & 16.8 & 77.0 & 86.5 & 51.1 & 69.7 & 64.8 & 54.1 & 69.7 & 63.5 & 96.6 & 67.1 & 77.7 & 72.1 & 87.1 & 84.5 \\
Cylinder3D~\cite{zhou2020cylinder3d} & L & 77.2 & 82.8 & 29.8 & 84.3 & 89.4 & 53.0 & 79.3 & 77.2 & 73.4 & 84.6 & 69.1 & 97.7 & 70.2 & 80.3 & 75.5 & 90.4 & 87.6 \\

CMDFusion~\cite{cen2023cmdfusion} & L & 80.8 & 83.5 & 45.7 & 94.5 & 91.4 & 76.7 & 87.0 & 77.2 & 73.0 & 85.6 & 77.3 & 97.4 & 69.2 & 79.5 & 75.5 & 91.0 & 88.5 \\
SPVCNN++~\cite{tang2020searching} & L & 81.1 & 86.4 & 43.1 & 91.9 & 92.2 & 75.9 & 75.7 & 83.4 & 77.3 & 86.8 & 77.4 & 97.7 & 71.2 & 81.1 & 77.2 & 91.7 & 89.0 \\
SVQNet~\cite{chen2023svqnet} & L & 81.3 & 84.5 & 41.8 & 93.4 & 92.5 & 69.2 & 85.5 & 83.7 & 78.4 & 84.5 & 77.5 & 97.2 & 70.4 & 81.7 & 77.9 & 91.8 & 90.2 \\
LidarMultiNet~\cite{ye2023lidarmultinet} & L & 81.4 & 80.4 & 48.4 & 84.3 & 90.0 & 71.5 & 87.2 & 85.2 & 80.4 & 86.9 & 74.8 & 97.8 & 67.3 & 80.7 & 76.5 & 92.1 & 89.6 \\
LiDARFormer~\cite{zhou2024lidarformer} & L & 81.5 & 84.4 & 40.8 & 84.7 & 92.6 & 72.7 & 91.0 & 84.9 & 81.7 & 88.6 & 73.8 & 97.9 & 69.3 & 81.7 & 77.4 & 92.4 & 89.6 \\
SphereFormer~\cite{lai2023spherical} & L & 81.9 & 83.3 & 39.2 & 94.7 & 92.5 & 77.5 & 84.2 & 84.4 & 79.1 & 88.4 & 78.3 & 97.9 & 69.0 & 81.5 & 77.2 & 93.4 & 90.2 \\
\midrule
2DPASS~\cite{yan20222dpass} & LC & 80.8 & 81.7 & 55.3 & 92.0 & 91.8 & 73.3 & 86.5 & 78.5 & 72.5 & 84.7 & 75.5 & 97.6 & 69.1 & 79.9 & 75.5 & 90.2 & 88.0 \\
PMF~\cite{zhuang2021perception} & LC & 77.0 & 82.1 & 40.3 & 80.9 & 86.4 & 63.7 & 79.2 & 79.8 & 75.9 & 81.2 & 67.1 & 97.3 & 67.7 & 78.1 & 74.5 & 89.9 & 88.5 \\
2D3DNet~\cite{genova2021learning} & LC & 80.0 & 83.0 & 59.4 & 88.0 & 85.1 & 63.7 & 84.4 & 82.0 & 76.0 & 84.8 & 71.9 & 96.9 & 67.4 & 79.8 & 76.0 & 92.1 & 89.2 \\
MSeg3D~\cite{li2023mseg3d} & LC & 81.1 & 83.1 & 42.5 & 94.9 & 92.0 & 67.1 & 78.6 & 85.7 & 80.5 & 87.5 & 77.3 & 97.7 & 69.8 & 81.2 & 77.8 & 92.4 & 90.1 \\
U2MKD~\cite{sun2024uni}& LC & 84.2 & 86.0 & 67.3 & 93.0 & 92.1 & 79.0 & 89.3 & 84.8 & 80.1 & 87.8 & 77.0 & 97.8 & 70.6 & 81.5 & 78.0 & 93.1 & 90.7 \\
TASeg~\cite{wu2024taseg} & LC & 84.6 & 87.1 & 69.4 & 90.5 & 92.2 & 78.7 & 90.4 & 86.3 & 81.9 & 88.3 & 75.9 & 97.8 & 70.9 & 81.0 & 78.2 & 93.4 & 91.2 \\
\midrule
\rowcolor{lightgray}
UniD-shift(ours) & LC & 81.2 & 83.4 & 61.9 & 92.0 & 86.7 & 73.0 & 85.7 & 81.4 & 76.1 & 87.4 & 68.4 & 97.7 & 67.6 & 80.6 & 76.6 & 92.3 & 89.3 \\

\bottomrule
\end{tabular}
}
\begin{minipage}{0.98\textwidth}\footnotesize
\textbf{Notes.} \textbf{L}: LiDAR-only. \textbf{LC}: LiDAR-Camera fusion. 
\end{minipage}
\end{table*}

\section{Failure Cases}
\label{Failure Cases}

\begin{figure}[h]
    \centering
    \includegraphics[width=\linewidth]{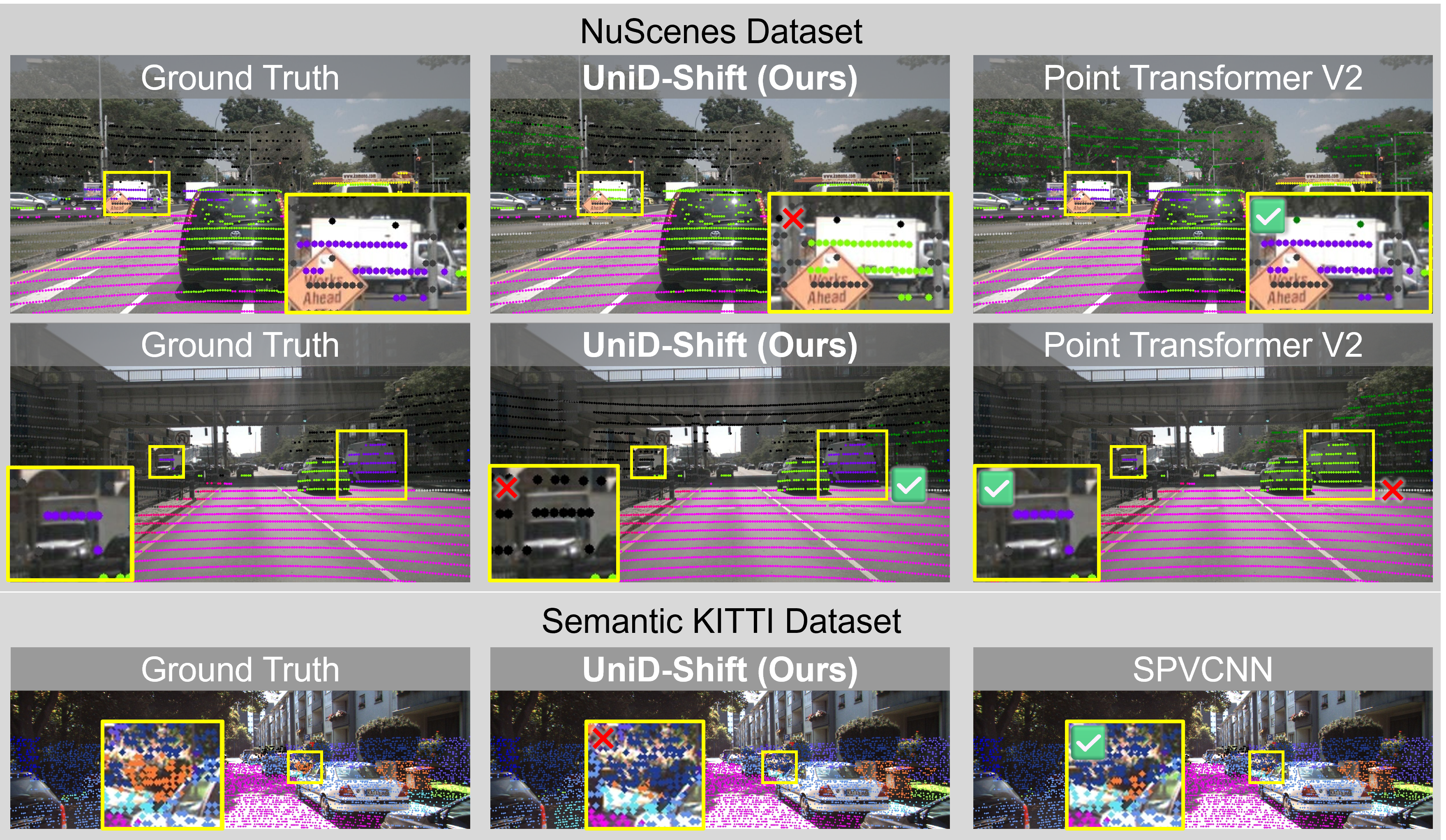}
    \caption{Representative failure cases on the nuScenes and Semantic KITTI validation datasets. }
    \label{fig:failure_cases}
\end{figure}

Figure~\ref{fig:failure_cases} illustrates representative cases in which the proposed framework encounters difficulties with distant and low-resolution targets. In these situations, the projected image features become severely compressed, and the corresponding point cloud samples contain only a few valid returns. Limited geometric evidence and reduced appearance detail create conditions in which the fused representation loses part of the fine structural patterns. Models that operate solely on point clouds occasionally retain these distant instances because their predictions rely more directly on sparse geometric clusters, whereas the multimodal fusion tends to emphasize regions with stronger cross-view correspondence.

The wider scene layout remains stable in the majority of samples, and the predicted masks maintain coherent spatial organization even when these local errors appear. The observed behavior indicates that the remaining challenges concentrate on long-range perception and extremely sparse regions. These conditions highlight potential directions for refinement, including improved handling of distant depth intervals and more adaptive treatment of sparse samples within the fusion process.

\begin{figure}[h]
    \centering
    \includegraphics[width=0.46\textwidth]{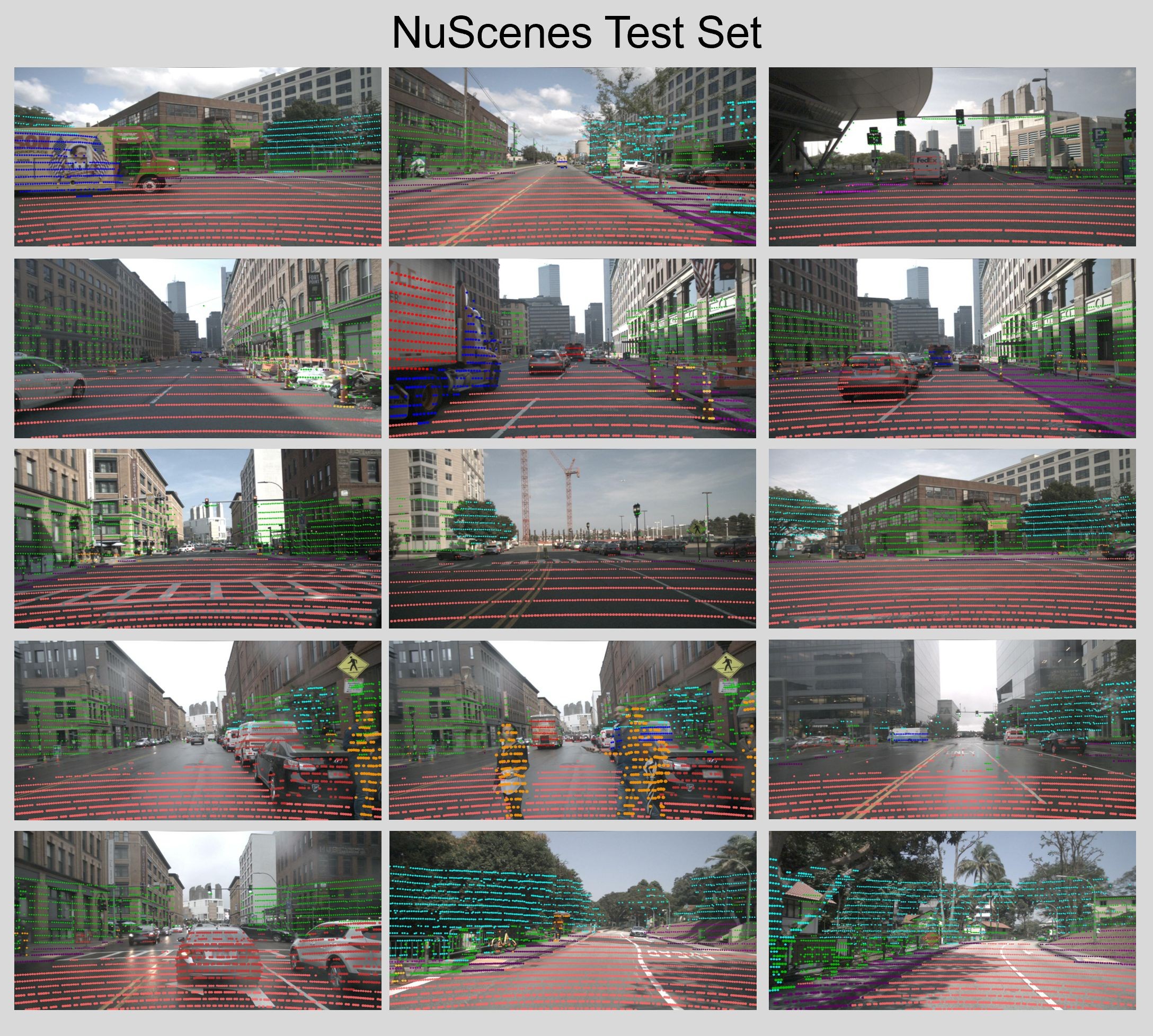}
    \vspace{-8pt}
    \caption{More visualization of object segmentation performance on the Nuscenes Test set.}
    \vspace{-8pt}
    \label{fig:nu_ki_test}
\end{figure}

\begin{figure}[h]
    \centering
    \includegraphics[width=0.47\textwidth]{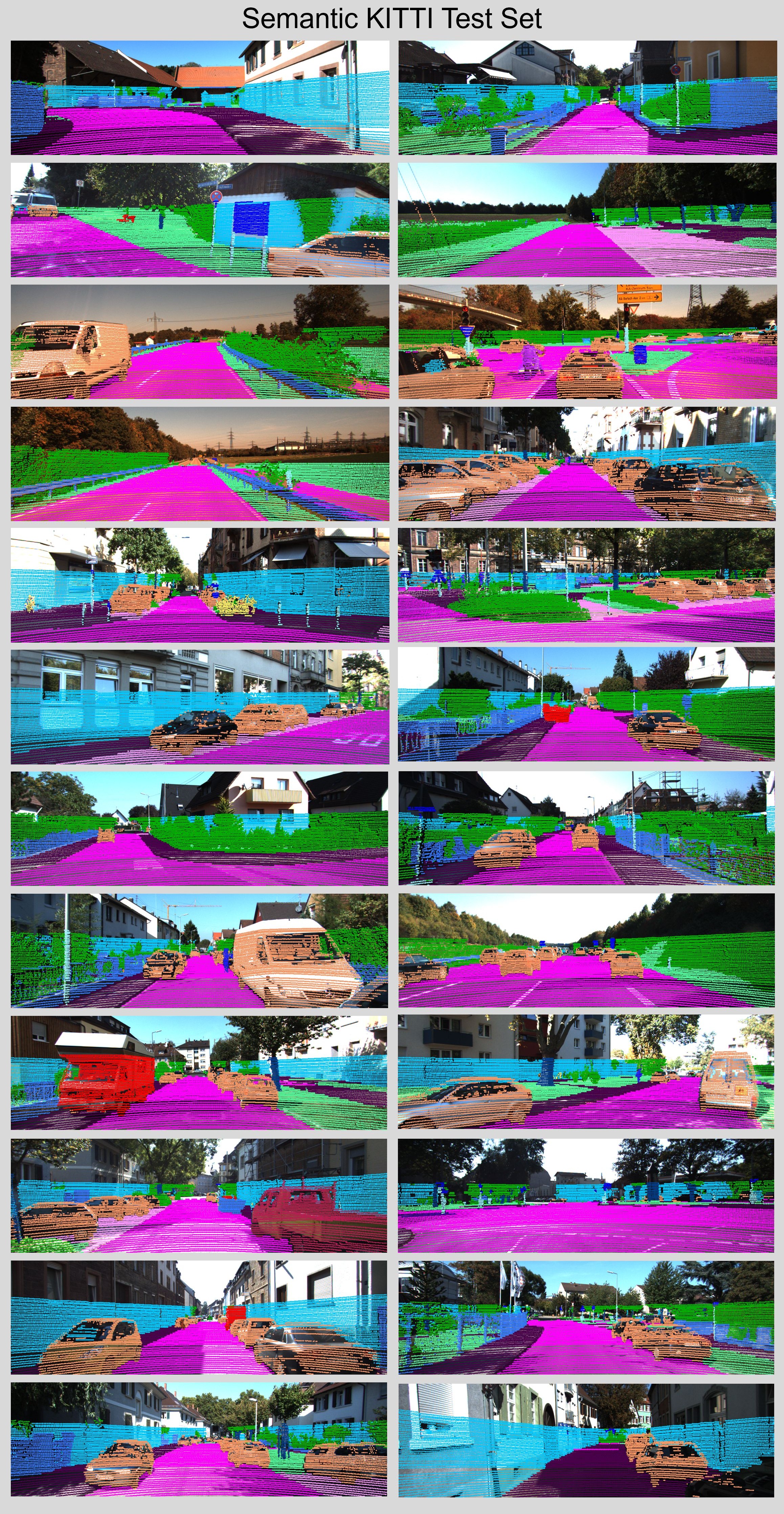}
    \vspace{-8pt}
    \caption{More visualization of object segmentation performance on the SemanticKitti Test set.}
    \vspace{-8pt}
    \label{fig:se_ki_test}
\end{figure}

\section{More Visualization}
\label{More Visualization}

\begin{figure*}[h]
    \centering
    \includegraphics[width=1\textwidth]{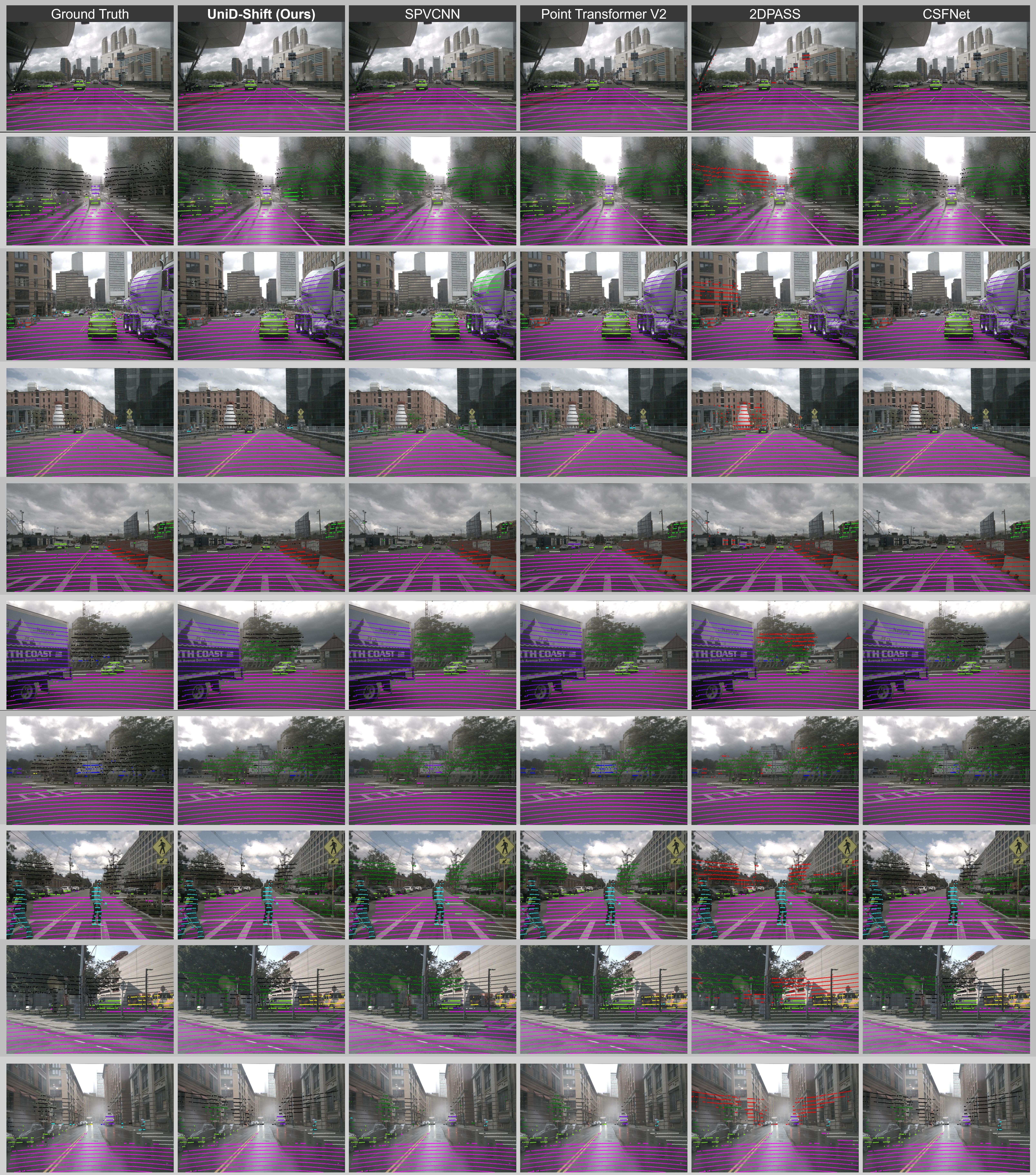}
    \vspace{-8pt}
    \caption{More visualizations comparing object segmentation performance with other methods on the nuScenes validation set.}

    \vspace{-8pt}
    \label{fig:More_vis_nu}
\end{figure*}

\begin{figure*}[h]
    \centering
    \includegraphics[width=1\textwidth]{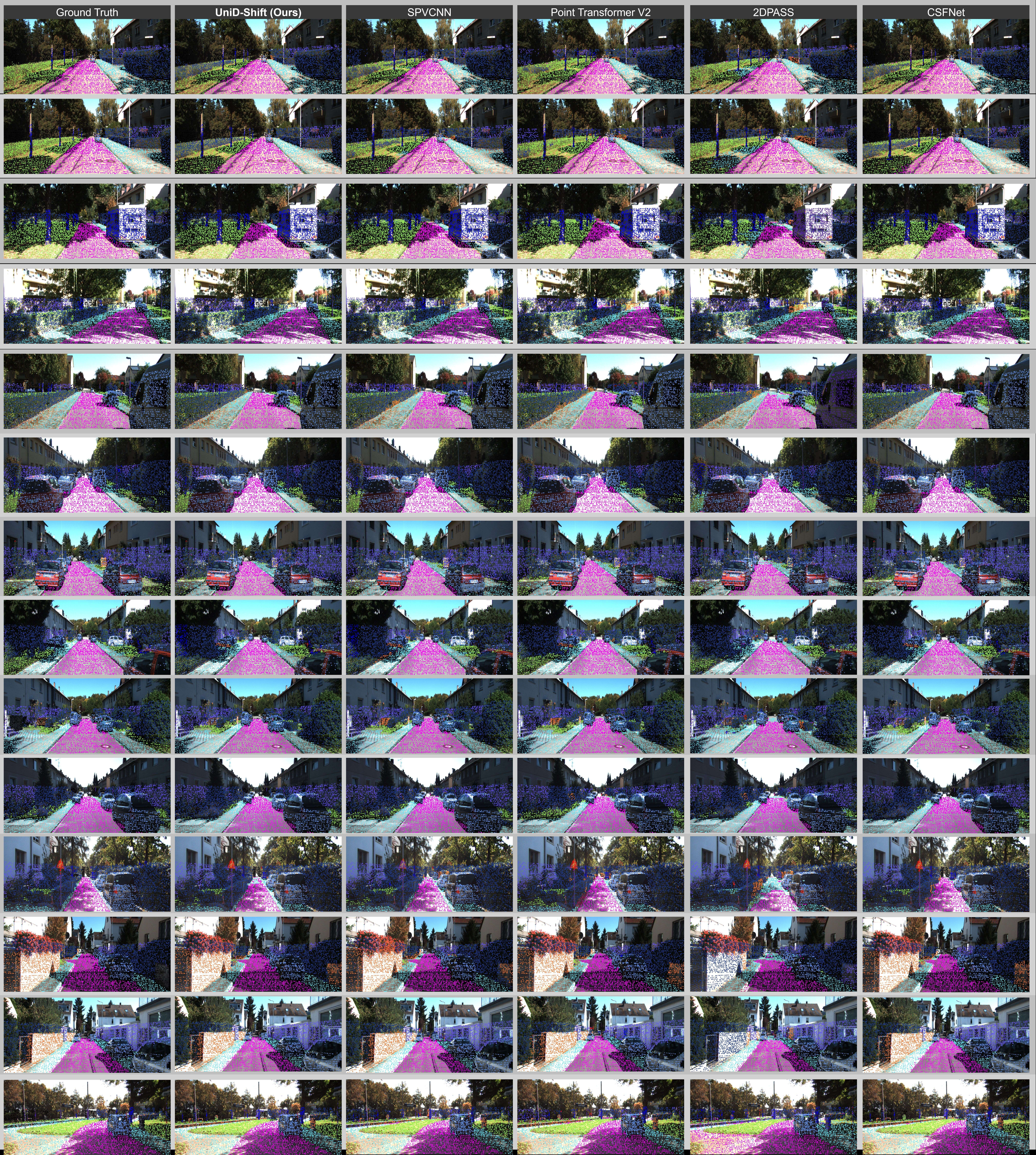}
    \vspace{-8pt}
    \caption{More visualizations comparing object segmentation performance with other methods on the SemanticKitti validation set.}

    \vspace{-8pt}
    \label{fig:More_vis_SK}
\end{figure*}

Additional visualizations are provided in Fig.~\ref{fig:More_vis_nu} and Fig.~\ref{fig:More_vis_SK}, offering extended visual evidence of the segmentation of UniD-Shift over a broad range of outdoor environments.

\end{document}